\documentclass[10pt,journal,compsoc]{IEEEtran}

\ifCLASSOPTIONcompsoc
  \usepackage[nocompress]{cite}
\else
  \usepackage{cite}
\fi

\ifCLASSINFOpdf
\else
\fi

\usepackage{bm}
\usepackage{amsmath}
\usepackage{amssymb}
\usepackage{multirow}
\usepackage{subfigure}
\usepackage{array}
\usepackage{amsthm}
\usepackage{amsfonts}
\usepackage{graphicx}
\usepackage{rotating}

\newtheorem{definition}{Definition}
\newcommand{\citet}{\cite}
\newcommand{\figref}[1]{Figure \ref{#1}}
\newcommand{\tabref}[1]{Table \ref{#1}}
\newcommand{\secref}[1]{Section \ref{#1}}
\newcommand{\appendixref}[1]{Appendix}
\newcommand{\equref}[1]{Equation (\ref{#1})}

\begin{document}
%
\title{Heterogeneous Graph Representation Learning with Relation Awareness}
%
%
%

\author{Le~Yu,
        Leilei~Sun,
        Bowen~Du,
        Chuanren~Liu,
        Weifeng~Lv,
        Hui~Xiong,~\IEEEmembership{Fellow,~IEEE}
\IEEEcompsocitemizethanks{\IEEEcompsocthanksitem L. Yu, L. Sun, B. Du and W. Lv are with the SKLSDE and BDBC Lab, Beihang University, Beijing, 100191, China.\protect\\
E-mail: yule@buaa.edu.cn, leileisun@buaa.edu.cn, dubowen@buaa.edu.cn, lwf@buaa.edu.cn
\IEEEcompsocthanksitem C. Liu is with the Department of Business Analytics and Statistics, the University of Tennessee, Knoxville, TN 37996, USA.\protect\\
E-mail: cliu89@utk.edu
\IEEEcompsocthanksitem H. Xiong is with the Hong Kong University of Science and Technology (Guangzhou) 
Thrust of Artificial Intelligence Nansha, Guangzhou, 511400, Guangdong, China and the Hong Kong University of Science and Technology, Department of Computer Science and Engineering, Hong Kong SAR, China.\protect\\
E-mail: xionghui@gmail.com}

\thanks{(Corresponding author: Leilei Sun.)}}

%
%

\markboth{IEEE TRANSACTIONS ON KNOWLEDGE AND DATA ENGINEERING,~Vol.~XX, No.~X, XX~XXXX}%
{Yu \MakeLowercase{\textit{et al.}}: Heterogeneous Graph Representation Learning with Relation Awareness}
%

\IEEEtitleabstractindextext{%
\begin{abstract}
Representation learning on heterogeneous graphs aims to obtain meaningful node representations to facilitate various downstream tasks, such as node classification and link prediction. 
Existing heterogeneous graph learning methods are primarily developed by following the propagation mechanism of node representations. 
There are few efforts on studying the role of relations for improving the learning of more fine-grained node representations. 
Indeed, it is important to collaboratively learn the semantic representations of relations and discern node representations with respect to different relation types. 
To this end, in this paper, we propose a \textbf{R}elation-aware \textbf{H}eterogeneous \textbf{G}raph \textbf{N}eural \textbf{N}etwork, namely R-HGNN, to learn node representations on heterogeneous graphs at a fine-grained level by considering relation-aware characteristics. 
Specifically, a dedicated graph convolution component is first designed to learn unique node representations from each relation-specific graph separately. 
Then, a cross-relation message passing module is developed to improve the interactions of node representations across different relations. 
Also, the relation representations are learned in a layer-wise manner to capture relation semantics, which are used to guide the node representation learning process. 
Moreover, a semantic fusing module is presented to aggregate relation-aware node representations into a compact representation with the learned relation representations. 
Finally, we conduct extensive experiments on a variety of graph learning tasks, and experimental results demonstrate that our approach consistently outperforms existing methods among all the tasks.
\end{abstract}

}

\maketitle

\IEEEdisplaynontitleabstractindextext

%
\IEEEpeerreviewmaketitle

\IEEEraisesectionheading{\section{Introduction}\label{section-1}}
\IEEEPARstart{H}{eterogeneous} graphs are pervasive in real-world scenarios, such as academic networks, e-commerce and social networks \cite{DBLP:journals/sigkdd/SunH12,DBLP:journals/tkde/ShiLZSY17}.
Learning rich information in heterogeneous graphs such as meaningful node representations could facilitate various tasks, including node classification \cite{DBLP:conf/kdd/DongCS17,DBLP:conf/www/ZhangXKLMZ18}, node clustering \cite{DBLP:conf/aaai/LiKRY19}, link prediction \cite{DBLP:conf/icdm/DongTWTCRC12,DBLP:conf/aaai/LiSCLTL20} and item recommendation \cite{DBLP:journals/tkde/ShiHZY19}.

\begin{figure}[!htbp]
\centering
\includegraphics[width=1.0\columnwidth]{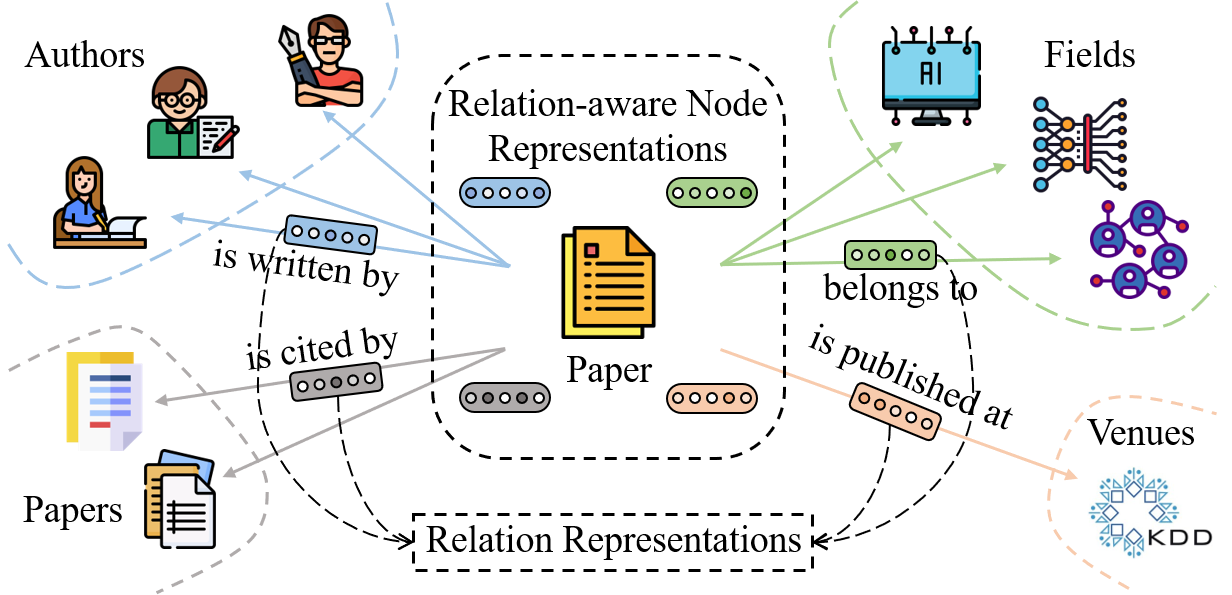}
\caption{
Motivation illustration. For a target node (see the central paper node), it is necessary to explicitly learn disparate representations of the target node with respect to different relations. Moreover, it is also important to learn the representations of relations and fuse the relation-aware representations of target nodes semantically.
}
\label{fig:motivation}
\end{figure}

In heterogeneous graphs, nodes are usually connected with different types of neighbors via different types of relations. Take the heterogeneous academic graph in \figref{fig:motivation} as an example, a target paper node is connected with nodes of authors, papers, fields and venues via "is written by", "is cited by", "belongs to" and "is published at" relations, respectively. Different types of relations can reflect disparate characteristics of the target nodes. For instance, the "belongs to" relation often reveals the paper's research topic, and the "is published at" relation tends to indicate the paper's technical quality. Therefore, it is essential to explicitly capture the underlying semantics of relations and learn \textit{relation-aware} node representations by maintaining a group of relation-specific node representations to encode more fine-grained information. The learned relation-aware node representations can reflect node characteristics with regard to each specified relation and make adaptive contributions of different relations for various downstream tasks. When estimating the popularity of papers, the "belongs to" relation that conveys the trending research topics would be more important. When we want to infer the paper category, the "is published at" relation may help more because it reflects whether a paper is more relevant to theoretical analysis or applied sciences.

However, this task is challenging in how to design a framework to ponder on the roles of nodes and relations in heterogeneous graphs in a collaborative manner. 
We summarize the relevant existing methods in the following three major ways. 
The first way falls into applying Graph Neural Networks (GNNs) for representation learning on graphs.
Recent GNNs, including GCN \cite{DBLP:conf/iclr/KipfW17}, GraphSAGE \cite{DBLP:conf/nips/HamiltonYL17} and GAT \cite{DBLP:conf/iclr/VelickovicCCRLB18}, have shown the superiority in modelling graph-structured data.
However, most GNNs were designed for homogeneous graphs with only one type of nodes and one type of edges, and they cannot directly handle different types of nodes and relations in heterogeneous graphs.

The second way focuses on designing specialized GNNs to learn node representations in heterogeneous graphs. \citet{DBLP:conf/www/WangJSWYCY19} designed HAN by leveraging pre-defined meta-paths and the attention mechanism to learn on heterogeneous graphs. \citet{DBLP:conf/kdd/ZhangSHSC19} presented HetGNN to consider the heterogeneity of node features and neighbors using Bi-LSTMs. \citet{DBLP:conf/aaai/HongGLYLY20} designed type-aware attention layers in HetSANN to study on neighboring nodes and associated edges with different types. \citet{DBLP:conf/www/HuDWS20} introduced HGT to investigate heterogeneous graphs using type-specific parameters based on the Transformer \cite{DBLP:conf/nips/VaswaniSPUJGKP17}. Although these methods provide some insights on heterogeneous graph learning, they primarily capture the characteristics of nodes. The relation semantics, which are also essential, have not been explicitly studied yet.

The third way is founded on modelling the properties of relations, which also carry essential information in graphs. RGCN \cite{DBLP:conf/esws/SchlichtkrullKB18} was proposed to deal with multiple relations in knowledge graphs. \citet{DBLP:conf/icdm/ZhuZPZW19} designed RSHN to learn on the constructed edge-centric coarsened line graph to tackle the relation diversity. \citet{DBLP:conf/aaai/LuSH019} presented RHINE to handle the affiliation and interaction relations. \citet{DBLP:conf/kdd/CenZZYZ019} introduced GATNE to model different types of relations between users and items. \citet{DBLP:conf/sigir/JinG0JL20} proposed MBGCN to capture multi-typed user behaviors by embedding propagation layers. While promising, these relation-centered methods still fail to explicitly learn relation semantics and the features of nodes with different types are not well discriminated.

To this end, we propose a \textbf{R}elation-aware \textbf{H}eterogeneous \textbf{G}raph \textbf{N}eural \textbf{N}etwork (R-HGNN), to learn not only fine-grained node representations according to different types of relations, but also the semantic representations of relations. Specifically, we first design a graph convolution module to propagate information on each relation-specific graph separately and learn node representation specified to the corresponding relation. Then, we present a cross-relation message passing module to improve the interactions of node representations across different relations. Next, the semantic representations of relations are explicitly learned layer by layer to guide the node representation learning process. Finally, to facilitate downstream tasks, the relation-aware node representations are semantically aggregated into a compact representation based on the learned relation representations. Extensive experiments are conducted on various graph tasks and the results show that our approach outperforms existing methods consistently among all the tasks. 
Our key contributions include:
\begin{itemize}
    \item 
    We propose a relation-aware node representation learning method. For each node, we derive a fine-grained representation from a group of relation-specific node representations, where each relation-specific representation reflects the characteristics of the node with regard to a specified relation.
    
    
    \item Accompanied with the relation-aware node representation learning process, we also provide a parallel relation representation learning module to learn the semantic representations of relations and guide the node representation learning process collaboratively.
    

    \item A semantic fusing module is proposed to aggregate relation-aware node representations into a compact representation to facilitate downstream tasks, considering the semantic characteristics of relations.
\end{itemize}

The rest of this paper is organized as follows:
\secref{section-2} summarizes previous research related to the studied problem.
\secref{section-3} formalizes the studied problem.
\secref{section-4} presents the framework and introduces each component of our model.
\secref{section-5} evaluates the proposed model through experiments.
Finally, \secref{section-6} concludes the entire paper.

\section{Related work}\label{section-2}
This section reviews existing literature related to our work, and also points out the differences of previous studies with our research.

\textbf{Graph Mining.}
Over the past decades, a great number of efforts have been made on graph mining. Classical methods based on manifold learning mainly focus on reconstructing graphs, such as Locally Linear Embedding (LLE) \cite{roweis2000nonlinear} and Laplacian Eigenmaps (LE) \cite{DBLP:conf/nips/BelkinN01}. Inspired by the Skip-gram model \cite{DBLP:conf/nips/MikolovSCCD13}, more advanced methods were proposed to learn representations of nodes in the network, including DeepWalk \cite{DBLP:conf/kdd/PerozziAS14}, node2vec \cite{DBLP:conf/kdd/GroverL16} and metapath2vec \cite{DBLP:conf/kdd/DongCS17}. These methods first adopt random walk strategy to generate sequences of nodes and then use Skip-gram to maximize co-occurrence probability of nodes in the same sequence.
However, the above methods only studied on the graph topology structure and could not consider node attributes, resulting in inferior performance. These methods are outperformed by the recently proposed GNNs, which could handle node attributes and the graph structure simultaneously.

\textbf{Graph Neural Networks.}
Recent years have witnessed the success of applying GNNs in various applications, such as node classification \cite{DBLP:conf/iclr/KipfW17,DBLP:conf/nips/HamiltonYL17}, graph classification \cite{DBLP:conf/iclr/XuHLJ19}, traffic prediction \cite{DBLP:conf/ijcai/YuYZ18}, and recommendation systems \cite{DBLP:conf/kdd/YingHCEHL18,DBLP:conf/sigir/Wang0WFC19,DBLP:conf/sigir/0001DWLZ020}. GNNs first propagate information among nodes and their neighbors, and then provide node representations by aggregating the received information. Generally, GNNs could be divided into spectral-based and spatial-based methods. As a spectral-based method, GCN \cite{DBLP:conf/iclr/KipfW17} introduces a localized first-order approximation and performs convolution in the Fourier domain. As spatial-based methods, GraphSAGE \cite{DBLP:conf/nips/HamiltonYL17} propagates information in the graph domain directly and utilizes different functions to aggregate neighbors' information. GAT \cite{DBLP:conf/iclr/VelickovicCCRLB18} leverages the attention mechanism to adaptively select more important neighbors. 
However, most existing GNNs were designed for homogeneous graphs, and could not handle different types of nodes and relations in heterogeneous graphs.

\textbf{Relational Graph Learning.}
There are some attempts to investigate the relations in graphs in recent years. \citet{DBLP:conf/esws/SchlichtkrullKB18} presented RGCN to model the relations in knowledge graphs by employing specialized transformation matrices for each relation type. \citet{DBLP:conf/icdm/ZhuZPZW19} designed RSHN to handle various relations by first building edge-centric coarsened line graph to describe relation semantic information and then using the learned relation representations to aggregate neighboring nodes. \citet{DBLP:conf/aaai/LuSH019} first conducted empirical studies to divide relations into two categories, i.e., the affiliation relations and the interaction relations, and then introduced RHINE to deal with these relations. \citet{DBLP:conf/kdd/CenZZYZ019} proposed GATNE to capture the multi-type interactions between users and items, which could support both transductive and inductive learning. 
In the field of recommendation systems, several studies focused on the multi-behavior recommendation problem where multiple behaviors of users could be seen as different relations between users and items. \citet{DBLP:conf/cikm/ZhangMCX20} introduced MGNN to learn both behavior-shared and behavior-specific embeddings for users and items to model the collective effects of multi-typed behaviors. \citet{DBLP:conf/sigir/JinG0JL20} first constructed a graph to represent multi-behavior data and then designed MBGCN to learn the strength as well as semantics of behaviors by embedding propagation layers. \citet{DBLP:conf/aaai/XiaHXDZYPB21} proposed KHGT to study on knowledge-aware multi-behavior graph, which captured both type-specific user-item interactive patterns and cross-type behavior dependencies. Although the above methods focused on the modelling of relations, the semantic representations of relations are still not explicitly learned. Moreover, the features associated with different types of nodes in heterogeneous graphs are not well discriminated.

\textbf{Heterogeneous Graph Learning.} 
Recently, a number of efforts aimed to design GNNs for heterogeneous graphs learning.
\citet{DBLP:conf/www/WangJSWYCY19} presented HAN to learn the importance of neighbors and multiple hand-designed meta-paths based on an attention mechanism. \citet{DBLP:conf/www/0004ZMK20} considered the intermediate nodes in meta-paths and proposed MAGNN to aggregate the intra-meta-path and inter-meta-path information.
HetGNN \cite{DBLP:conf/kdd/ZhangSHSC19} first adopts a random walk strategy to sample neighbors and then uses specialized Bi-LSTMs to integrate heterogeneous node features and neighboring nodes. \citet{DBLP:conf/aaai/HongGLYLY20} designed HetSANN to learn on different types of neighboring nodes as well as the associated edges through type-aware attention layers, which could directly encode the graph information via a dedicated attention mechanism. 
Based on the architecture of Transformer \cite{DBLP:conf/nips/VaswaniSPUJGKP17}, \citet{DBLP:conf/www/HuDWS20} introduced HGT to learn the characteristics of different nodes and relations with type-specific parameters.
However, the above methods are mostly developed by following the propagation mechanism of node representations, and the role of relations has not been comprehensively exploited yet.

Different with the above mentioned methods, we consider the role of relations to improve the learning of more fine-grained node representations in heterogeneous graph learning. In particular, our approach could collaboratively learn both relation-aware node representations and semantic representations of relations.

\section{Preliminaries}
\label{section-3}

This section provides the definitions of heterogeneous graphs as well as the formalization of the studied problem. 

\subsection{Definitions}
\begin{definition}
    \textbf{Heterogeneous Graph}. A heterogeneous graph is defined as $\mathcal{G}=\left(\mathcal{V},\mathcal{E},\mathcal{A},\mathcal{R}\right)$ with a node type mapping function $\phi : \mathcal{V} \rightarrow \mathcal{A}$ and an edge type mapping function $\psi : \mathcal{E} \rightarrow \mathcal{R}$, where $\mathcal{V}$, $\mathcal{E}$, $\mathcal{A}$ and $\mathcal{R}$ correspond to the set of nodes, edges, node types and edge types, respectively.
    Each node $v \in \mathcal{V}$ and each edge $e \in \mathcal{E}$ belong to one specific type in $\mathcal{A}$ and $\mathcal{R}$, i.e., $\phi(v) \in \mathcal{A}$ and $\psi(e) \in \mathcal{R}$. Each heterogeneous graph has multiple node or edge types such that $|\mathcal{A}| + |\mathcal{R}| > 2$.
\end{definition}

\textbf{Example.} As shown in \figref{fig:motivation}, a heterogeneous academic graph contains multiple types of nodes (i.e., papers, authors, fields and venues) and relations (e.g., "is written by" relation between papers and authors, "belongs to" relation between papers and fields).

\begin{definition}
\textbf{Relation}.
A relation represents the connecting pattern of the source node, the corresponding edge and the target node.
Specifically, for an edge $e=(u,v)$ linked from source node $u$ to target node $v$, the corresponding relation is denoted by $\left \langle \phi(u), \psi(e), \phi(v) \right \rangle$. 
Naturally, the inverse relation is represented by $\left \langle \phi(v), \psi(e)^{-1}, \phi(u) \right \rangle$ in this paper.

\textbf{Example.} In \figref{fig:motivation}, the relations consist of "is written by", "is cited by", "belongs to" and "is published at". In this paper, we study the role of relations and learn relation-aware node representations.
\end{definition}

\subsection{Problem Formalization}
Given a heterogeneous graph $\mathcal{G}=(\mathcal{V},\mathcal{E},\mathcal{A},\mathcal{R})$, representation learning on heterogeneous graph aims to learn a function $f: \mathcal{V} \rightarrow \mathbb{R}^d$ to embed each node into a $d$-dimensional representation with $d \ll |V|$.
The learned representations should capture both node features and relation information to facilitate various tasks, such as node classification, node clustering and link prediction.

\section{Methodology}\label{section-4}
This section first presents the framework of the proposed model and then introduces each component step by step.

\subsection{Framework of the Proposed Model}
\begin{figure*}[!htbp]
    \centering
    \includegraphics[width=2.0\columnwidth]{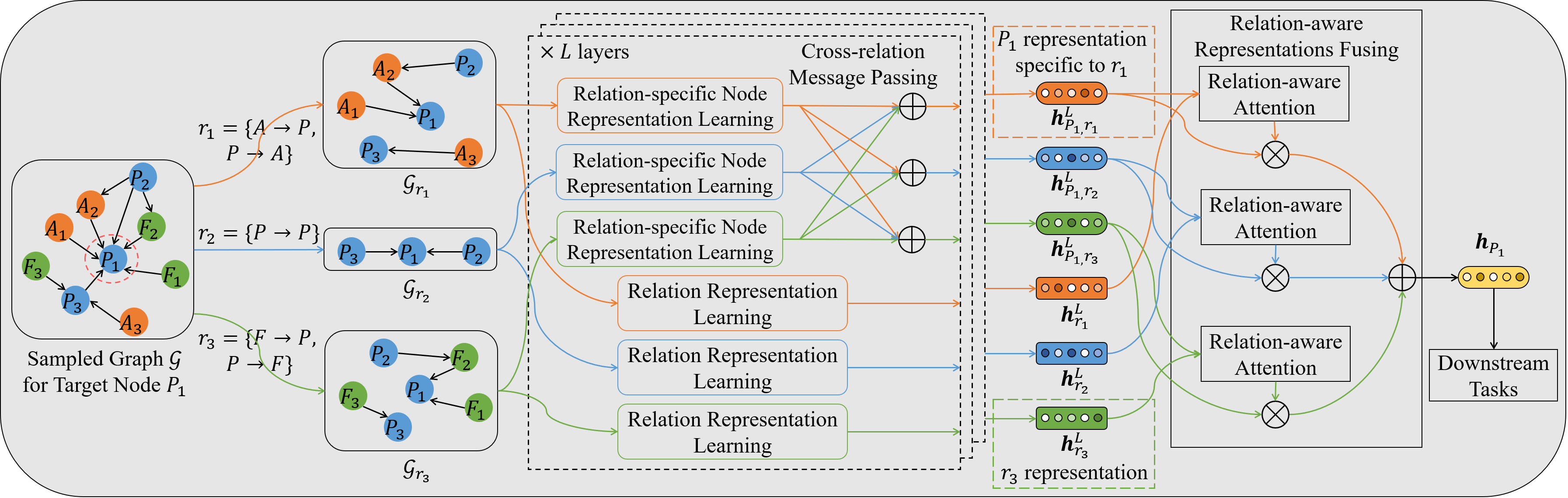}    
    \caption{Framework of the proposed model. R-HGNN could collaboratively learn relation-aware node representations for target node $P_1$ (i.e., $\bm{h}_{P_1,r_1}^L$, $\bm{h}_{P_1,r_2}^L$ and $\bm{h}_{P_1,r_3}^L$) as well as the relation representations for $r_1$, $r_2$ and $r_3$ (i.e., $\bm{h}_{r_1}^L$, $\bm{h}_{r_2}^L$ and $\bm{h}_{r_3}^L$). Finally, a compact representation $\bm{h}_{P_1}$ for node $P_1$ is provided to facilitate downstream tasks.}
    \label{fig:framework}
\end{figure*}

The framework of the proposed model is shown in \figref{fig:framework}, which takes a sampled graph $\mathcal{G}$ for the target node with node feature matrices as the input and provides the low-dimensional node representation $\bm{h}_v$ for $v \in \mathcal{V}$ as the output, which could be applied in various downstream tasks.

The proposed model consists of four components: relation-specific node representation learning, cross-relation message passing, relation representation learning and relation-aware representations fusing. 
Originally, the sampled heterogeneous graph for the target node is decomposed into multiple relation-specific graphs based on the types of relations.
The first component performs graph convolution to learn unique node representations from each relation-specific graph separately. 
The second component establishes connections to improve the interactions of node representations across different relations. 
The third component explicitly learns relation representation in a layer-wise manner, and uses them to guide the node representation learning process.
The fourth component aims to aggregate relation-aware node presentations into a compact representation considering the semantic characteristics of relations to facilitate downstream tasks, such as node classification, node clustering, and link prediction.

\subsection{Relation-specific Node Representation Learning}\label{section-3-relation_specific_node_representation_laerning}
As shown in \figref{fig:motivation}, in heterogeneous graphs, a target node is usually associated with multiple relations.
Existing heterogeneous graph learning methods are primarily designed by following the propagation mechanism of node representations, while the role of relations is not explicitly exploited. 
Therefore, we aim to learn node representations considering the relation-aware characteristics, which indicates that each node is associated with a relation-specific representation to reflect the characteristics of the node with regard to the corresponding relation.

To learn the relation-specific node representation for a target node, we first decompose the heterogeneous graph $\mathcal{G}$ into multiple relation-specific graphs $\left\{ \mathcal{G}_r, r \in \mathcal{R}\right\}$ based on the relation types.
Note that the inverse relation $r^{-1}$ is also added in graph $\mathcal{G}_r$ to allow the two connected nodes propagate information from each other. 
Then, we design a dedicated graph convolution module to learn unique node representations from each relation-specific graph. Finally, we present a weighted residual connection to combine the target node features and the aggregated neighboring information adaptively. Details of the two modules are introduced as follows.

\begin{figure}[!htbp]
    \centering
    \includegraphics[width=0.98\columnwidth]{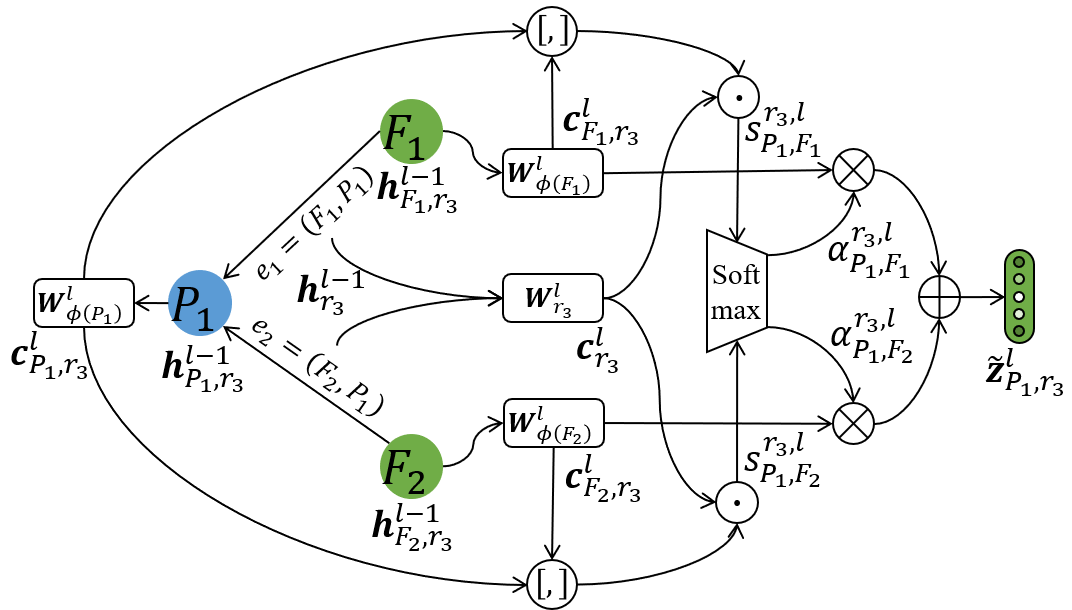}
    \caption{Illustration of the convolution on relation-specific graph $\mathcal{G}_{r_3}$. Activation functions are omitted for simplicity.}
    \label{fig:single_relation_convolution}
\end{figure}

\textbf{Relation-specific Convolution.} 
The process of the convolution on each relation-specific graph is shown in \figref{fig:single_relation_convolution}.
Mathematically, we first project source node $u$, target node $v$ and relation $\psi(e)$\footnote{Since each edge type is unique in heterogeneous graphs, we use edge type $\psi(e)$ to denote the relation $\left \langle \phi(u), \psi(e), \phi(v) \right \rangle$ for simplicity, unless otherwise stated.} into their latent spaces via the node-type and relation-type specific transformation matrices via the following equations,
\begin{equation}
    \bm{c}_{u,{\psi(e)}}^l = \bm{W}_{\phi(u)}^l \bm{h}_{u,{\psi(e)}}^{l-1},
\end{equation}
\begin{equation}
    \bm{c}_{v,\psi(e)}^l = \bm{W}_{\phi(v)}^l \bm{h}_{v,\psi(e)}^{l-1},
\end{equation}
\begin{equation}
    \bm{c}_{\psi(e)}^l = \bm{W}_{\psi(e)}^l \bm{h}_{\psi(e)}^{l-1},
\end{equation}
where $\bm{W}_{\phi(u)}^l$, $\bm{W}_{\phi(v)}^l$ and $\bm{W}_{\psi(e)}^l$ are type-specific trainable transformation matrices for source node $u$, target node $v$ and relation $\psi(e)$ at the $l$-th layer, respectively. 
$\bm{h}_{u,\psi(e)}^{l-1}$, $\bm{h}_{v,{\psi(e)}}^{l-1}$ and $\bm{h}_{\psi(e)}^{l-1}$ are the representations of source node $u$, target node $v$ and relation $\psi(e)$ in the corresponding relation-specific graph at layer $l-1$. We set $\bm{h}_{u,\psi(e)}^0$, $\bm{h}_{v,\psi(e)}^0$ and $\bm{h}_{\psi(e)}^0$ to their original features $\bm{x}_u$, $\bm{x}_v$ and $\bm{x}_{\psi(e)}$ initially. The original node features $\bm{x}_u$ and $\bm{x}_v$ are usually given by the graph. The original relation feature $\bm{x}_{\psi(e)}$ is represented in one-hot encoding, where the entry of 1 corresponds to the relation type. Then we calculate the normalized importance of source node $u$ to target node $v$ by
\begin{equation}
    s_{v,u}^{\psi(e),l} = LeakyReLU\left({\bm{c}_{\psi(e)}^l}^\top \left[\bm{c}_{v,\psi(e)}^l, \bm{c}_{u,{\psi(e)}}^l \right]\right),
\end{equation}
\begin{equation}
    \alpha_{v,u}^{\psi(e),l} = \frac{\exp{\left(s_{v,u}^{\psi(e),l}\right)}}{\sum_{u\prime \in \mathcal{N}_{\psi(e)}(v)} \exp{\left(s_{v,u\prime}^{\psi(e),l}\right)}},
\end{equation}
where $\left[\cdot,\cdot\right]$ is the concatenation operation, $\mathcal{N}_{\psi(e)}(v)$ denotes the set of $v$'s neighbors with relation $\psi(e)$.
Finally, the information of $v$'s neighbors with relation $\psi(e)$ is aggregated with the learned importance as follows,
\begin{equation}
    \widetilde{\bm{z}}_{v,\psi(e)}^l = ReLU\left(\sum_{u \in \mathcal{N}_{\psi(e)}(v)} \alpha_{v,u}^{\psi(e),l} \cdot \bm{c}_{u,{\psi(e)}}^l\right).
\end{equation}
It is worth mentioning that previous research design trainable parameters in each layer separately to aggregate neighbor information, which may decrease the correlations of layers. To improve the layer relevance, we leverage the relation representations, which are propagated in a layer-wise manner (elaborate latter in \secref{section-3-relation_propagate}), to calculate the importance of target node's neighbors. Since the relation representations are propagated layer by layer, the interactions between adjacent layers could be enhanced.

\textbf{Weighted Residual Connection.} In addition to aggregating neighbor information by the relation-specific convolution, the features of the target node are also assumed to be important, because they reflect the node properties inherently. However, simply incorporating the target node with and neighbor information via summation could not distinguish their different importance. 

Hence, we combine the target node features and the aggregated neighbor information via a residual connection \cite{DBLP:conf/cvpr/HeZRS16} with trainable weight parameters by
\begin{equation}
    \bm{z}_{v,\psi(e)}^l = \lambda_{\phi(v)}^l \cdot \widetilde{\bm{z}}_{v,\psi(e)}^l + \left(1-\lambda_{\phi(v)}^l\right) \cdot \bm{W}_{\phi(v),align}^l \bm{h}_{v,\psi(e)}^{l-1},
\end{equation}
where $\lambda_{\phi(v)}^l$ controls the combination importance and is trainable during the training process. $\bm{W}_{\phi(v),align}^l$ is used to align the dimensions of $\widetilde{\bm{z}}_{v,\psi(e)}^l$ and $\bm{h}_{v,\psi(e)}^{l-1}$. Due to the design of trainable importance, our model could aggregate the target features and neighbor information adaptively.

\subsection{Cross-relation Message Passing}
\label{section-3-cross_relation_message_passing}
Through the above relation-specific node representation learning component, we could obtain multiple representations of the target node which are specific to relation types. In fact, different relations that the target node interacts with tend to be correlated with each other. Treating the node representations under each relation disparately would neglect the relation interactions and lead to inferior performance, which is empirically validated in \secref{section-4-ablation_study}. Therefore, it is necessary to propagate the messages across different relations to provide more informative node representations. However, naive pooling operations across relations become infeasible because they fail to discern the node representations with regard to different relation types. In this paper, we establish connections of node representations to improve message passing across different relations and automatically distinguish the importance of relations. 

Formally, let $\mathcal{R}(v)$ denote the set of relations that node $v$ associated with. Given the learned node representations of $v$ with respect to each relation, i.e., $\left\{\bm{z}_{v,r}^l, r \in \mathcal{R}(v)\right\}$, the message passing between relation $\psi(e)$ and the relations in $\mathcal{R}(v)$ is implemented by
\begin{equation}
    \bm{h}_{v,\psi(e)}^l = \sum_{r \in \mathcal{R}(v)} \beta_{\psi(e),r}^l \cdot \bm{z}_{v,r}^l,
\end{equation}
where $\beta_{\psi(e),r}^l$ is the normalized relevance of relation $r$ to relation $\psi(e)$ at the $l$-th layer, and it is calculated by
\begin{equation}
    \beta_{\psi(e),r}^l = \frac{\exp{\left(LeakyReLU\left({\bm{q}_{\psi(e)}^l}^\top\bm{z}_{v,r}^l\right)\right)}}{\sum_{r\prime \in \mathcal{R}(v)} \exp{\left(LeakyReLU\left({\bm{q}_{\psi(e)}^l}^\top\bm{z}_{v,r\prime}^l\right)\right)}}.
\end{equation}
$\bm{q}_{\psi(e)}^l$ is the trainable attention vector specific to relation $\psi(e)$ at the $l$-th layer, which is used to control the information flow between relation $\psi(e)$ and the relations in $\mathcal{R}(v)$. By establishing the connections of node representations across relations, each unique representation specific to the corresponding relation could contact with other relations and become more informative.

\subsection{Relation Representation Learning} \label{section-3-relation_propagate}
As illustrated in \secref{section-1}, relations reflect the connecting patterns of nodes and carry rich information. Therefore, in addition to learning relation-aware node representations by a specialized node representation propagation mechanism through the above two components, our approach also founds on the semantic characteristics of relations, which are rarely studied in existing methods. It is worth noticing that although existing methods like HAN, RGCN and HGT design trainable parameters specific to meta-paths or relations to capture such characteristics, they fail to focus on the role of relations explicitly, which indicates that the semantic representations of different relations are ignored.

To explicitly learn the relation semantic representations, we propose a general propagation mechanism for relation representations, which could be formalized as 
\begin{equation}
    \bm{h}_{\psi(e)}^l = PROPAGATE^l(\bm{h}_{\psi(e)}^{l-1}, \bm{extra}).
    \label{equ:general_propagate_relation}
\end{equation}
$PROPAGATE^l(\cdot)$ represents the propagation mechanism to update relation representations at layer $l$, which could be achieved by a succinct linear propagation or a sophisticated gated updating propagation \cite{DBLP:journals/neco/HochreiterS97}. $\bm{extra}$ denotes the inputs except for the relation representation $\bm{h}_{\psi(e)}^{l-1}$, such as the hidden states of relations in the gated updating propagation. In this work, we implement the relation propagating mechanism without considering $\bm{extra}$ and rewrite \equref{equ:general_propagate_relation} as follows,
\begin{equation}
    \bm{h}_{\psi(e)}^l = \bm{W}_{\psi(e),upd}^l \bm{h}_{\psi(e)}^{l-1} + \bm{b}_{\psi(e),upd}^l,
\end{equation}
where $\bm{W}_{\psi(e),upd}^l$ and $\bm{b}_{\psi(e),upd}^l$ are the trainable parameters to update the representations for relation $\psi(e)$ at layer $l$. The relation representations could not only capture the semantic characteristics of relations, but also guide the learning process of node representations, making nodes and relations to be collaboratively learned (introduced in \secref{section-3-relation_specific_node_representation_laerning}).

\subsection{Relation-aware Representations Fusing}
We define that a R-HGNN layer is composed of the aforementioned three components and stack $L$ layers to receive information from multi-hop neighbors. Finally, the $L$ layers could provide relation-aware node representations for target node $v$, i.e., $\left\{ \bm{h}_{v,r}^L, r \in \mathcal{R}(v)\right\}$, as well as the representations of relations associated with $v$, that is, $\left\{ \bm{h}_{r}^L, r \in \mathcal{R}(v)\right\}$. For downstream tasks, a compact node representation is usually required. One could apply simple pooling operations (e.g., average or max pooling) on the relation-aware representations to obtain the compact representation, but such operations fail to consider the importance of node representations across different relations (empirically validated in \secref{section-4-ablation_study}). Therefore, we design a semantic fusing component to aggregate the relation-aware node representations into a compact node representation to facilitate various downstream tasks. 

In particular, this component takes $\left\{ \bm{h}_{v,r}^L, r \in \mathcal{R}(v)\right\}$ and $\left\{ \bm{h}_{r}^L, r \in \mathcal{R}(v)\right\}$ as inputs, and provides a compact representation $\bm{h}_v$ for node $v$ by a relation-aware attention mechanism via the following equations,
\begin{equation}
    \gamma_{v,r} = \frac{\exp{\left(LeakyReLU\left(\left({\bm{V}_{r}\bm{h}_{v,r}^L}\right)^\top\bm{E}_{r}\bm{h}_{r}^L\right)\right)}}{\sum_{r^\prime \in \mathcal{R}(v)}\exp{\left(LeakyReLU\left(\left({\bm{V}_{r^\prime}\bm{h}_{v,r^\prime}^L}\right)^\top\bm{E}_{r^\prime}\bm{h}_{r^\prime}^L\right)\right)}},
\end{equation}
\begin{equation}
    \bm{h}_{v} = \sum_{r \in \mathcal{R}(v)}\gamma_{v,r} \cdot \bm{V}_{r}\bm{h}_{v,r}^L,
\end{equation}
where $\gamma_{v,r}$ represents the learned importance of relation $r$ to node representation $\bm{h}_{v}$. $\bm{V}_{r}$ and $\bm{E}_{r}$ represent the transformation matrix for node representation $\bm{h}_{v,r}^L$ and relation representation $\bm{h}_{r}^L$, respectively. Finally, the compact node representation $\bm{h}_{v}$ are obtained through the weighted aggregation using the learned relation importance, which could be applied in downstream tasks. It is worth noticing that the relation-aware node representations are semantically aggregated by considering the relation representations, and such a design is in line with the motivation of our approach, that is, investigating on the role of relations to improve the learning of node representations.

\subsection{End-to-End Learning Process}
We build the proposed R-HGNN by first stacking $L$ R-HGNN layers to learn relation-aware node representations and then employing the relation-aware representations fusing component to integrate multiple representations component into a compact representation. We also adopt the multi-head attention mechanism to enhance the model capacity and make the training process more stable, where the outputs of different heads are combined via the concatenation operation. The proposed R-HGNN could be trained in an end-to-end manner with the following strategies.

\textbf{Semi-supervised learning strategy}. For tasks where the labels are available (e.g., node classification), R-HGNN could be optimized by minimizing the following cross entropy loss,
\begin{equation}
    \label{equ:semi_supervised_loss}
    \mathcal{L} = - \sum_{v \in \mathcal{V}_{label}} \sum_{c=1}^{C} y_{v,c} \cdot \log \hat{y}_{v,c},
\end{equation}
where $\mathcal{V}_{label}$ denotes the set of labeled nodes. $y_{v,c}$ and $\hat{y}_{v,c}$ represent the ground truth and predicted possibility of node $v$ at the $c$-th dimension. $\hat{y}_{v,c}$ can be obtained from a classifier (e.g., a single-layer neural network), which takes $\bm{h}_v$ as the input and provides $\hat{\bm{y}}_v$ as the output.

\textbf{Unsupervised learning strategy}. For tasks without using node labels (e.g., link prediction), R-HGNN could be optimized by minimizing the binary cross entropy with negative sampling strategy in Skip-gram \cite{DBLP:conf/nips/MikolovSCCD13}, which takes representations of paired nodes as the inputs,
\begin{equation}
    \label{equ:unsupervised_loss}
    \mathcal{L} = - \sum_{(v, u) \in \Omega_P} \log \sigma \left(\bm{h}_v^\top \bm{h}_u\right) - \sum_{(v^\prime, u^\prime) \in \Omega_N} \log \sigma \left(-\bm{h}_{v^\prime}^\top \bm{h}_{u^\prime}\right),
\end{equation}
where $\sigma(\cdot)$ is the sigmoid activation function, $\Omega_P$ and $\Omega_N$ denote the set of positive observed and negative sampled node pairs, respectively.

\subsection{Analysis of the Proposed Model}
Here we give the complexity analysis and summarize the advantages of R-HGNN as follows.

\textbf{Model Complexity}. 
Our R-HGNN is highly efficient and could be easily parallelized. 
Let $N_{in}^l$ and $N_{out}^l$ denote the input and output dimension of node representations at the $l$-th layer. Let $R_{in}^l$ and $R_{out}^l$ represent the input and output dimension of relation representations at the $l$-th layer. Let $L$ denote the number of stacked R-HGNN layers and $d$ denote the dimension of final compact node representation. The time complexity of calculating the relation-specific node representation with respect to relation $r \in \mathcal{R}$ is linear to the number of nodes and edges in the relational graph $\mathcal{G}_r$. It can be denoted by $\mathcal{O}(\alpha_1 |\mathcal{V}_r| + \beta |\mathcal{E}_r| + \gamma)$, where $|\mathcal{V}_r|$ and $|\mathcal{E}_r|$ are the number of nodes and edges in the relational graph $\mathcal{G}_r$. $\alpha_1=N_{out}^l(N_{in}^l + |\mathcal{R}|)$, $\beta=R_{in}^l N_{out}^l$ and $\gamma=R_{in}^l R_{out}^l$. Hence, the relation-specific node representations can be calculated individually across nodes and relations. When fusing the relation-aware representations to provide the compact node representation, the time complexity is linear to the number of nodes in the heterogeneous graph $\mathcal{G}$, which can be represented as $\mathcal{O}(\alpha_2 |\mathcal{V}_r|)$ with $\alpha_2=d|\mathcal{R}|(R_{out}^L  + N_{out}^L)$.

\textbf{Model Superiority}.
1) R-HGNN can directly learn on the heterogeneous graph by first decomposing the original graph into several relation-specific graphs naturally and then learning node representations via the dedicated node representation learning component. Due to the exploration on the inherent structure of heterogeneous graph, our model leverages the information of all the nodes and achieves more comprehensive node representations.
2) R-HGNN discerns node representations with respect to different relation types, where each relation-specific representation reflects the node disparate characteristics. R-HGNN also explicitly captures the role of relations through learning relation semantics, which are used to guide the learning process of relation-aware node representations. The design of learning relation-aware node representations improves the model learning ability and captures more fine-grained information. 
3) R-HGNN has the advantage of discovering more important relations, which is beneficial for heterogeneous graph analysis. According to the learned importance of each relation-specific node representation, we could intuitively observe which relation makes more contributions for the downstream task and analyze the results better.

\section{Experiments}\label{section-5}
This section evaluates the performance of the proposed method by experiments on various graph learning tasks, including node classification, node clustering, node visualization and link prediction.

\begin{table*}[!htbp]
\centering
\caption{Statistics of the datasets.}
\label{tab:dataset_description}
\resizebox{2\columnwidth}{!}
{
\begin{tabular}{c|c|c|c|c|c|c}
\hline
Datasets     & Nodes                                                                                                                                       & Edges                                                                                                                 & Features                                                              & Feature Extraction                                                                                                                   & Split Strategy                                                      & Split Sets                                                                                 \\ \hline
IMDB         & \begin{tabular}[c]{@{}c@{}}\# Movie (M): 4,076\\ \# Director (D): 1,999\\ \# Actor (A): 5,069\end{tabular}                                  & \begin{tabular}[c]{@{}c@{}}\# M-D: 4,076\\ \# M-A: 12,228\end{tabular}                                                & \begin{tabular}[c]{@{}c@{}}M:1,537\\ D:1,537\\ A:1,537\end{tabular}   & \begin{tabular}[c]{@{}c@{}}M:bag-of-words of keywords\\ D:average of directed movies\\ A:average of acted movies\end{tabular} & \begin{tabular}[c]{@{}c@{}}Random Split\\ (following \cite{DBLP:conf/www/WangJSWYCY19})\end{tabular}     & \begin{tabular}[c]{@{}c@{}}Train: 817\\ Validation: 407\\ Test: 2,852\end{tabular}         \\ \hline
OGB-MAG      & \begin{tabular}[c]{@{}c@{}}\# Paper (P): 736,389\\ \# Author (A): 1,134,649\\ \# Field (F): 59,965\\ \# Institution (I): 8,740\end{tabular} & \begin{tabular}[c]{@{}c@{}}\# P-A: 7,145,660\\ \# P-P: 5,416,271\\ \# P-F: 7,505,078\\ \# A-I: 1,043,998\end{tabular} & \begin{tabular}[c]{@{}c@{}}P:256\\ A:128\\ F:128\\ I:128\end{tabular} & \begin{tabular}[c]{@{}c@{}}P:Word2Vec \& metapath2vec\\ A:metapath2vec\\ F:metapath2vec\\ I:metapath2vec\end{tabular}         & \begin{tabular}[c]{@{}c@{}}Time-based Split\\ (following \cite{DBLP:conf/nips/HuFZDRLCL20})\end{tabular} & \begin{tabular}[c]{@{}c@{}}Train: 629,571\\ Validation: 64,879\\ Test: 41,939\end{tabular} \\ \hline
OAG-Venue    & \begin{tabular}[c]{@{}c@{}}\# Paper (P): 166,065\\ \# Author (A): 510,189\\ \# Field (F): 45,717\\ \# Institution (I): 9,079\end{tabular}   & \begin{tabular}[c]{@{}c@{}}\# P-A: 477,676\\ \# P-P: 851,644\\ \# P-F: 1,700,497\\ \# A-I: 612,872\end{tabular}       & \begin{tabular}[c]{@{}c@{}}P:768\\ A:768\\ F:400\\ I:400\end{tabular} & \begin{tabular}[c]{@{}c@{}}P:XLNet\\ A:average of published papers\\ F:metapath2vec\\ I:metapath2vec\end{tabular}               & \begin{tabular}[c]{@{}c@{}}Time-based Split\\ (following \cite{DBLP:conf/www/HuDWS20})\end{tabular} & \begin{tabular}[c]{@{}c@{}}Train: 106,058\\ Validation: 24,255\\ Test: 35,752\end{tabular} \\ \hline
OAG-L1-Field & \begin{tabular}[c]{@{}c@{}}\# Paper (P): 119,483\\ \# Author (A): 510,189\\ \# Venue (V): 6,934\\ \# Institution (I): 9,079\end{tabular}    & \begin{tabular}[c]{@{}c@{}}\# P-A: 340,959\\ \# P-P: 329,703\\ \# P-V: 119,483\\ \# A-I: 612,872\end{tabular}         & \begin{tabular}[c]{@{}c@{}}P:768\\ A:768\\ V:400\\ I:400\end{tabular} & \begin{tabular}[c]{@{}c@{}}P:XLNet\\ A:average of published papers\\ V:metapath2vec\\ I:metapath2vec\end{tabular}               & \begin{tabular}[c]{@{}c@{}}Time-based Split\\ (following \cite{DBLP:conf/www/HuDWS20})\end{tabular} & \begin{tabular}[c]{@{}c@{}}Train: 81,071\\ Validation: 16,439\\ Test: 21,973\end{tabular}  \\ \hline
\end{tabular}
}
\end{table*}

\subsection{Description of Datasets}
We conduct experiments on four real-world datesets, containing a small-scale dataset (IMDB) and three large-scale datasets (OGB-MAG, OAG-Venue and OAG-L1-Field).

\begin{itemize}
    \item \textbf{IMDB}\footnote{https://data.world/data-society/imdb-5000-movie-dataset}:
    Following \citet{DBLP:conf/www/WangJSWYCY19}, we extract a subset of IMDB and construct a heterogeneous graph containing movies (M), directors (D) and actors (A). The movies are divided into three categories: Action, Comedy and Drama. The movie features are denoted by the bag-of-words representation of the plot keywords. Director/actor features are the average representation of movies that they directed/acted. We use the random split strategy in \cite{DBLP:conf/www/WangJSWYCY19} to spilt the dataset.
    
    \item \textbf{OGB-MAG}:
    OGB-MAG \cite{DBLP:conf/nips/HuFZDRLCL20} is a heterogeneous academic network extracted from the Microsoft Academic Graph (MAG), consisting of paper (P), authors (A), fields (F) and institutions (I). Papers are published on 349 different venues. Each paper is associated with a Word2Vec feature. All the other types of nodes are not associated with input features and we adopt the metapath2vec \cite{DBLP:conf/kdd/DongCS17} model to generate their features. We use the time-based split strategy in \cite{DBLP:conf/nips/HuFZDRLCL20} to conduct experiments.
    
    \item \textbf{OAG-Venue}:
    OAG-Venue \cite{DBLP:conf/www/HuDWS20} is a heterogeneous graph in the Computer Science (CS) domain, which consists of paper (P), authors (A), fields (F) and institutions (I). The papers are published on 241 different venues. Paper's features are obtained from a pre-trained XLNet \cite{DBLP:conf/nips/YangDYCSL19} and the feature of each author is the average of his/her published paper representations. The features of other types of nodes are generated by the metapath2vec \cite{DBLP:conf/kdd/DongCS17} model. The dataset is split by the time-based split strategy in \cite{DBLP:conf/www/HuDWS20}.
    
    \item \textbf{OAG-L1-Field}:
    OAG-L1-Field \cite{DBLP:conf/www/HuDWS20} is another heterogeneous graph in the CS domain, containing papers (P), authors (A), venues (V) and institutions (I). The papers belong to 52 different $L1$-level fields. The feature extraction and split strategy are the same with those used in OAG-Venue.
\end{itemize}
Statistics of the datasets are summarized in \tabref{tab:dataset_description}. 
Please refer to the \appendixref{section-appendix-datasets} for more details of the datasets.

\subsection{Compared Methods}
We compare with several state-of-the-art baselines, which could be divided into four groups:

\textbf{Graph Topology-Agnostic Methods:}
\begin{itemize}
	\item \textbf{MLP} uses the multilayer perceptron to solely take node features as the input without considering the graph topology.
\end{itemize}

\textbf{Homogeneous Graph Learning Methods:}

\begin{itemize}
    \item \textbf{GCN}
    performs graph convolutions in the Fourier domain via leveraging the localized first-order approximation \cite{DBLP:conf/iclr/KipfW17}.
    
    \item \textbf{GraphSAGE}
    propagates information in the graph domain and designs different aggregate functions the aggregate neighbor's information. \cite{DBLP:conf/nips/HamiltonYL17}.
    
    \item \textbf{GAT}
    employs the attention mechanism to assign different importance to the neighbors adaptively \cite{DBLP:conf/iclr/VelickovicCCRLB18}. 
\end{itemize}

\textbf{Relational Graph Learning Methods:}
\begin{itemize}
    \item \textbf{RGCN}
    investigates on the relations in knowledge graphs by employing specialized transformation matrices for each type of relations \cite{DBLP:conf/esws/SchlichtkrullKB18}.
    
    \item \textbf{RSHN}
    first constructs edge-centric coarsened line graph to capture relation semantic information and then uses relation representations to aggregate neighboring nodes \cite{DBLP:conf/icdm/ZhuZPZW19}.
\end{itemize}

\textbf{Heterogeneous Graph Learning Methods:}
\begin{itemize}
    \item \textbf{HAN}
    leverages an attention mechanism to aggregate neighbor information in heterogeneous graphs via multiple manually designed meta-paths \cite{DBLP:conf/www/WangJSWYCY19}.
    
    \item \textbf{HetSANN}
    designs type-aware attention layers to obtain node representations with the consideration of different types of neighboring nodes as well as their associated edges \cite{DBLP:conf/aaai/HongGLYLY20}.
    
    \item \textbf{HGT}
    utilizes type-specific parameters to capture the characteristics of different nodes and relations inspired by the Transformer architecture \cite{DBLP:conf/www/HuDWS20}.
\end{itemize}

\subsection{Experimental Setup}
For models that require meta-paths as inputs, we use $MDM$ and $MAM$ as meta-paths on IMDB, use $PAP$, $PFP$ and $PPP$ as meta-paths on OGB-MAG and OAG-Venue, and use $PAP$, $PVP$ and $PPP$ as meta-paths on OAG-L1-Field.
Following \citet{DBLP:conf/www/WangJSWYCY19}, we test models designed for homogeneous graphs (i.e., GCN, GraphSAGE and GAT) on the graph generated by each meta-path and report the best performance.
We feed all the meta-paths into HAN as it could handle multiple meta-paths.
For RGCN, RSHN, HetSANN, HGT and the proposed R-HGNN, which could leverage the features of all types of nodes, we add a projection layer for each type of node to align the feature dimension.
All methods are optimized via the Adam \cite{DBLP:journals/corr/KingmaB14} optimizer with the cosine annealing learning rate scheduler \cite{DBLP:conf/iclr/LoshchilovH17}.
We use dropout \cite{DBLP:journals/jmlr/SrivastavaHKSS14} to prevent over-fitting, and apply grid search to select the best dropout and learning rate for all the methods (see more details in the \appendixref{section-appendix-hyper-parameters}). 
For all the methods, we search the hidden dimension of node representation in [32, 64, 128] and [128, 256, 512] on the small-scale and large-scale datasets. For methods that utilize the attention mechanism, including GAT, RSHN, HAN, HetSANN, HGT and our R-HGNN, we search the number of attention heads in [1, 2, 4, 8, 16]. Since R-HGNN additionally captures the characteristics of relations, we set the hidden dimension of relation representation to 64 on all the datasets. 
We implement the two-layer R-HGNN with PyTorch \cite{DBLP:conf/nips/PaszkeGMLBCKLGA19} and Deep Graph Library (DGL) \cite{DBLP:journals/corr/abs-1909-01315}.

We train all the methods in a full-batch manner on the small-scale dataset.
When training on large-scale datasets, it is intractable to train the GNN methods on the whole graph directly due to the memory usage. Hence, we adopt a neighbor sampling strategy to train GNN methods in a mini-batch manner. Specifically, for a target node, we sample a fixed number of neighbors per edge type at each layer uniformly and then make the node gather messages from the sampled neighbors layer by layer. We set the number of sampled neighbors per edge type to 10 in the first layer, and the number of sampled neighbors is increased by one layer by layer (e.g., 10, 11, $\cdots$). The sampling strategy is available in the DGL package\footnote{https://docs.dgl.ai/api/python/dgl.dataloading.html}. Note that HGT is equipped with a customized sampling method (HGSampling) for large-scale heterogeneous graphs, so we keep the specialized HGSampling for HGT in the experiments. HAN is not evaluated on large-scale datasets due to the difficulty in how to use sampling strategy to train HAN with the constraints of multiple mata-paths.
The inference of the GNN methods is performed on the whole graph with limited GPU memory via mini-batch and neighborhood sampling. This process is also provided by DGL\footnote{Please refer to https://docs.dgl.ai/en/latest/guide/minibatch-inference.html\#guide-minibatch-inference for more details of the implementation of inference process.}. We train all the methods with a fixed 200 epochs and use early stopping strategy with a patience of 50, which means the training process is terminated when the evaluation metrics on the validation set are not improved for 50 consecutive epochs. The model with the best performance on the validation set is used for testing. The codes and datasets are available at https://github.com/yule-BUAA/R-HGNN.

\begin{table*}[!htbp]
\centering
\caption{Comparisons with different methods on the node classification task.}
\label{tab:node_classification_results}
\begin{tabular}{c|c|cccccccccc}
\hline
Datasets                      & Metrics  & MLP    & GCN    & GraphSAGE & GAT    & RGCN   & RSHN   & HAN    & HetSANN & HGT    & R-HGNN         \\ \hline
\multirow{2}{*}{IMDB}         & Accuracy & 0.5547 & 0.6013 & 0.6287    & 0.6192 & 0.6339 & 0.6367 & 0.6318 & 0.6073  & 0.6364 & \textbf{0.6424} \\
                              & Macro-F1 & 0.5514 & 0.5975 & 0.6276    & 0.6197 & 0.6359 & 0.6337 & 0.6298 & 0.6069  & 0.6365 & \textbf{0.6417} \\ \hline
\multirow{2}{*}{OGB-MAG}      & Accuracy & 0.3243 & 0.3487 & 0.3704    & 0.3767 & 0.4796 & 0.4728 & ---    & 0.4781  & 0.4921 & \textbf{0.5204} \\
                              & Macro-F1 & 0.1426 & 0.1741 & 0.1885    & 0.1925 & 0.2766 & 0.2819 & ---    & 0.2821  & 0.3092 & \textbf{0.3206} \\ \hline
\multirow{2}{*}{OAG-Venue}    & Accuracy & 0.1082 & 0.1651 & 0.1762    & 0.1764 & 0.2397 & 0.2159 & ---    & 0.2581  & 0.2447 & \textbf{0.2887} \\
                              & Macro-F1 & 0.0548 & 0.1187 & 0.1148    & 0.1213 & 0.2099 & 0.1878 & ---    & 0.2257  & 0.2145 & \textbf{0.2585} \\ \hline
\multirow{2}{*}{OAG-L1-Field} & Accuracy & 0.3839 & 0.4915 & 0.5437    & 0.5531 & 0.5783 & 0.5738 & ---    & 0.5804  & 0.5713 & \textbf{0.5891} \\
                              & Macro-F1 & 0.1859 & 0.3408 & 0.3602    & 0.3770 & 0.3794 & 0.3656 & ---    & 0.3893  & 0.3657 & \textbf{0.4015} \\ \hline
\end{tabular}
\end{table*}

\begin{table*}[!htbp]
\centering
\caption{Comparisons with different methods on the node clustering task.}
\label{tab:node_clustering_results}
\begin{tabular}{c|c|cccccccccc}
\hline
Datasets                      & Metrics & MLP    & GCN    & GraphSAGE & GAT    & RGCN   & RSHN   & HAN    & HetSANN & HGT    & R-HGNN         \\ \hline
\multirow{2}{*}{IMDB}         & NMI     & 0.1837 & 0.2074 & 0.2551    & 0.2435 & 0.2605 & 0.2614 & 0.2503 & 0.2365  & 0.2592 & \textbf{0.2688} \\
                              & ARI     & 0.2067 & 0.2418 & 0.2957    & 0.2771 & 0.2953 & 0.3038 & 0.2856 & 0.2724  & 0.3003 & \textbf{0.3098} \\ \hline
\multirow{2}{*}{OGB-MAG}      & NMI     & 0.4951 & 0.4825 & 0.5226    & 0.5096 & 0.5814 & 0.5710 & ---    & 0.6600  & 0.6704 & \textbf{0.6744} \\
                              & ARI     & 0.4306 & 0.4372 & 0.4379    & 0.4244 & 0.4661 & 0.5022 & ---    & 0.5392  & 0.5459 & \textbf{0.5470} \\ \hline
\multirow{2}{*}{OAG-Venue}    & NMI     & 0.2118 & 0.2802 & 0.2789    & 0.2870 & 0.4593 & 0.4039 & ---    & 0.5135  & 0.5273 & \textbf{0.5390} \\
                              & ARI     & 0.2019 & 0.2511 & 0.2418    & 0.2614 & 0.4183 & 0.3797 & ---    & 0.4690  & 0.4847 & \textbf{0.4950} \\ \hline
\multirow{2}{*}{OAG-L1-Field} & NMI     & 0.1761 & 0.2913 & 0.3347    & 0.3189 & 0.3872 & 0.3491 & ---    & 0.2990  & 0.3533 & \textbf{0.4037} \\
                              & ARI     & 0.1554 & 0.2565 & 0.2801    & 0.2667 & 0.3411 & 0.2975 & ---    & 0.2748  & 0.2925 & \textbf{0.3619} \\ \hline
\end{tabular}
\end{table*}

\begin{figure*}[!htbp]
\centering
\includegraphics[width=2.0\columnwidth]{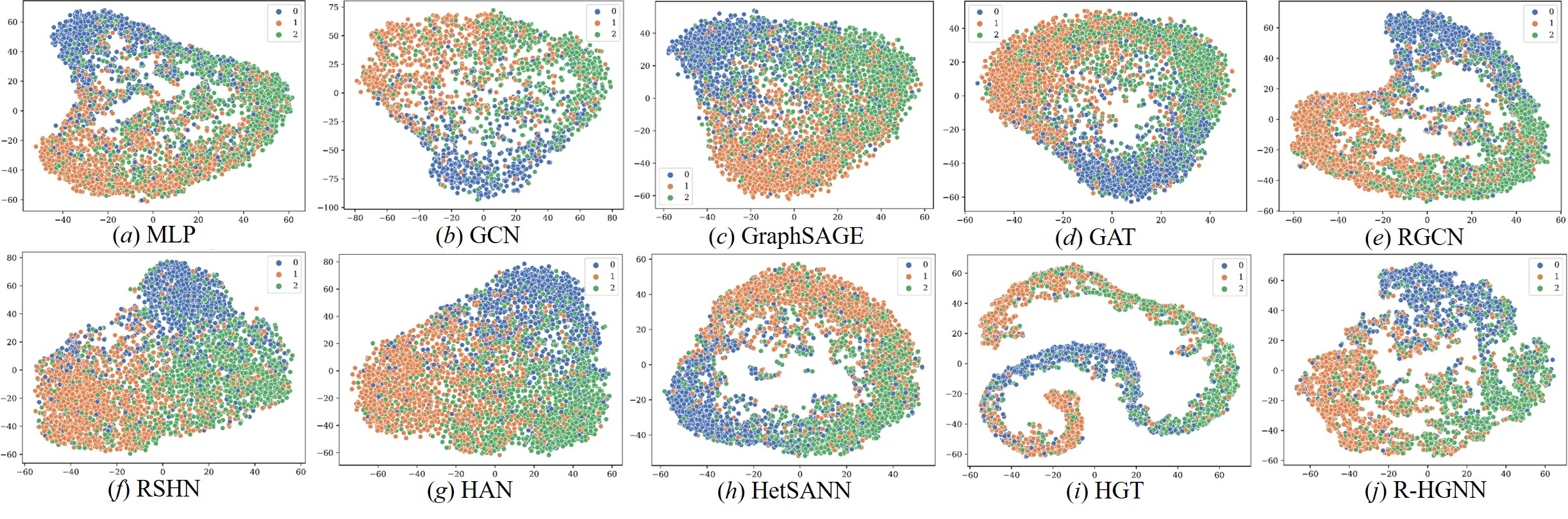}
\caption{
Visualization of the node representations on IMDB.
Each point indicates a movie and its color denotes movie category.}
\label{fig:node_visualization_IMDB}
\end{figure*}

\subsection{Node Classification}
We evaluate the model performance on the node classification task, which aims to predict the category of a movie (IMDB), the published venue of a paper (OGB-MAG and OAG-Venue) and the field that a paper belongs to (OAG-L1-Field). 

\textbf{Setting.} Following \citet{DBLP:conf/www/WangJSWYCY19}, we split the IMDB into training, validation and testing sets with the ratio of 2:1:7 randomly. Following \citet{DBLP:conf/nips/HuFZDRLCL20,DBLP:conf/www/HuDWS20}, the split strategy on OGB-MAG, OAG-Venue and OAG-L1-Field is based on the paper published years. Specifically, papers published before 2018, in 2018 and after 2018 are divided into training, validation and testing sets on OGB-MAG. Papers published before 2015, between 2015 and 2016, and after 2016 are divided into training, validation and testing sets on OAG-Venue and OAG-L1-Field. We feed the learned node representations into a linear classifier to get the final predictions. The objective function to be minimized is \equref{equ:semi_supervised_loss}. We use both Accuracy and Macro-F1 as evaluation metrics, and higher metrics correspond to better models. We run all the models for ten times and report the averaged performance in \tabref{tab:node_classification_results}.

\textbf{Result.} By analyzing the results in \tabref{tab:node_classification_results}, several conclusions could be summarized. Firstly, MLP performs worse than other methods, which indicates that leveraging the information of neighboring nodes could result in better performance. Secondly, homogeneous graph learning methods usually obtain worse performance than relation learning or heterogeneous graph learning methods, implying the necessity of leveraging the graph heterogeneity. Finally, R-HGNN outperforms all the baselines on all the datasets, and the improvements are even significant on complicated large-scale datasets. This observation demonstrates the superiority of R-HGNN in considering the role of relations, as well as learning relation-aware node representations. 

\subsection{Node Clustering}
The node clustering task is conducted to validate the effectiveness of the learned node representations. 

\textbf{Setting.} We first obtain node representations via the trained model and then feed the normalized representations into the k-means algorithm. Please refer to the \appendixref{section-appendix-node_clustering} for more details on the node clustering task. Normalized Mutual Information (NMI) and Adjusted Rand Index (ARI) are adopted as the evaluation metrics, and higher metrics correspond to better models. We run k-means for 10 times and report the average performance in \tabref{tab:node_clustering_results}.

\textbf{Result.}
From \tabref{tab:node_clustering_results}, we observe that R-HGNN performs better than all the baselines.
Moreover, methods designed for relation learning or heterogeneous graphs learning usually achieve better performance than homogeneous graph learning methods, demonstrating the importance of utilizing the graph heterogeneity.

\subsection{Node Visualization}
We conduct the node visualization task to provide a more intuitive comparison with R-HGNN and the baselines.

\textbf{Setting.} We visualize the movie nodes in IMDB into a low dimensional space. In detail, we project the learned representations of all the movie nodes into a 2-dimensional space using t-SNE \cite{maaten2008visualizing}. The result is shown in \figref{fig:node_visualization_IMDB}.

\textbf{Result.} From \figref{fig:node_visualization_IMDB}, we find that R-HGNN performs better than baselines on the node visualization task. Movies with the same category are gathered closely and the boundaries between movies with different categories are more obvious. The baselines either fail to gather movies with the same category together, or could not provide clear boundaries for movies with different categories.

\subsection{Link Prediction}
We also evaluate the effectiveness of different methods on the link prediction task.

\begin{table}[!htbp]
\centering
\caption{Performances on the link prediction task.}
\label{tab:link_prediction_results}
\resizebox{1.00\columnwidth}{!}{
\setlength{\tabcolsep}{1.0mm}
{
\begin{tabular}{c|c|c|ccccc}
\hline
Datasets                                                                    & Edges                & Metrics & RGCN   & RSHN   & HetSANN & HGT    & R-HGNN         \\ \hline
\multirow{4}{*}{\begin{tabular}[c]{@{}c@{}}OGB-\\ MAG\end{tabular}}         & \multirow{2}{*}{A-P} & RMSE    & 0.1506 & 0.1357 & 0.1218  & 0.1231 & \textbf{0.1178} \\
                                                                            &                      & MAE     & 0.0385 & 0.0364 & 0.0214  & 0.0292 & \textbf{0.0194} \\ \cline{2-8} 
                                                                            & \multirow{2}{*}{A-I} & RMSE    & 0.2387 & 0.1849 & 0.1422  & 0.1602 & \textbf{0.1129} \\
                                                                            &                      & MAE     & 0.0817 & 0.0607 & 0.0324  & 0.0437 & \textbf{0.0201} \\ \hline
\multirow{4}{*}{\begin{tabular}[c]{@{}c@{}}OAG-\\ Venue\end{tabular}}       & \multirow{2}{*}{A-P} & RMSE    & 0.2663 & 0.2416 & 0.2289  & 0.2349 & \textbf{0.2193} \\
                                                                            &                      & MAE     & 0.1021 & 0.1182 & 0.0767  & 0.0991 & \textbf{0.0683} \\ \cline{2-8} 
                                                                            & \multirow{2}{*}{A-I} & RMSE    & 0.3532 & 0.3437 & 0.3327  & 0.3304 & \textbf{0.3161} \\
                                                                            &                      & MAE     & 0.1817 & 0.2220 & 0.1975  & 0.2059 & \textbf{0.1793} \\ \hline
\multirow{4}{*}{\begin{tabular}[c]{@{}c@{}}OAG-\\ L1-\\ Field\end{tabular}} & \multirow{2}{*}{A-P} & RMSE    & 0.2428 & 0.2419 & 0.2041  & 0.2186 & \textbf{0.1886} \\
                                                                            &                      & MAE     & 0.0836 & 0.0979 & 0.0645  & 0.0831 & \textbf{0.0498} \\ \cline{2-8} 
                                                                            & \multirow{2}{*}{A-I} & RMSE    & 0.3360 & 0.3260 & 0.3181  & 0.3143 & \textbf{0.3089} \\
                                                                            &                      & MAE     & 0.1769 & 0.1946 & 0.1842  & 0.1838 & \textbf{0.1684} \\ \hline
\end{tabular}
}
}
\end{table}

\textbf{Setting.} We predict two types of links on large-scale datasets: composition between authors and papers (A-P) and affiliation between authors and institutions (A-I). We split the edges into training, validation and testing sets with the ratio of 3:1:1. Please see more details in the \appendixref{section-appendix-link_prediction}. The possibility that two nodes are connected by an edge is computed via the dot product of the node representations. The objective function to be minimized is \equref{equ:unsupervised_loss}. Root Mean Squared Error (RMSE) and Mean Absolute Error (MAE) are adopted as the evaluation metrics, and lower metrics correspond to better models. The performance is reported in \tabref{tab:link_prediction_results}.

\textbf{Result.} From \tabref{tab:link_prediction_results}, we could observe that our model outperforms all the other methods with a significant margin. It demonstrates that the proposed R-HGNN provides more effective node representations than baselines on the link prediction task.

\subsection{Ablation Study}\label{section-4-ablation_study}
We conduct the ablation study to validate the effectiveness of the components in R-HGNN on large-scale datasets for the node classification task. We investigate on the weighted residual connection, the cross-relation message passing and the relation-aware representations fusing components and design three variants, namely R-HGNN w/o WRC, R-HGNN w/o CMP, and R-HGNN w/o RRF. 

\begin{figure}[!htbp]
    \centering
    \includegraphics[width=\columnwidth]{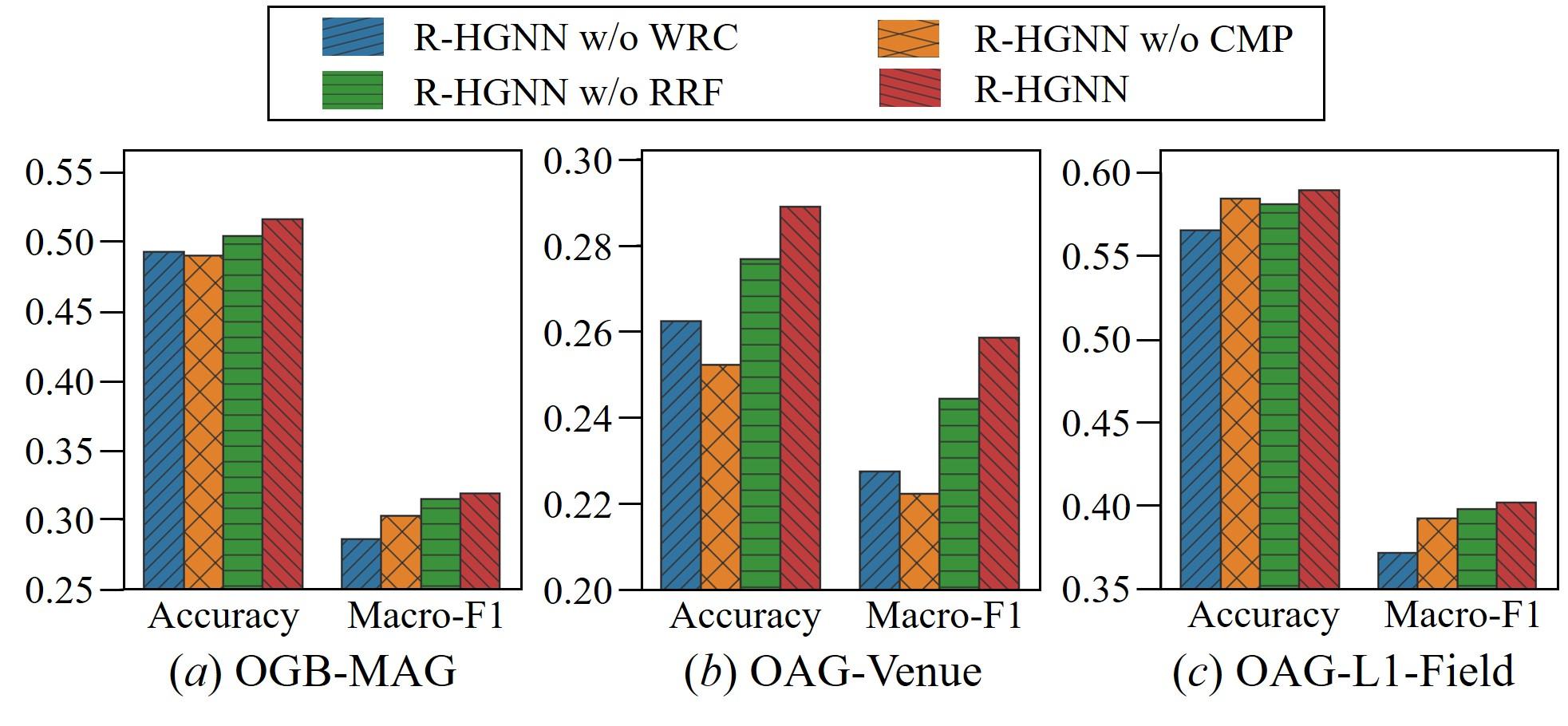}
\caption{
Effects of the WRC, CMP and RRF components on large-scale datasets for the node classification task.}
\label{fig:ablation_study}
\end{figure}

Specifically, R-HGNN w/o WRC removes the weighted residual connection module and does not take target node features into consideration. R-HGNN w/o CMP eliminates the connections of node representations across different relations, which means that node representations under different relations are independent and do not pass message with each other. R-HGNN w/o RRF replaces the relation-aware representations fusing module with the mean pooling operation, indicating that the relation-aware node representations are aggregated into a compact representation without distinguishing the importance of different relations.
Experimental results are shown in \figref{fig:ablation_study}.

From \figref{fig:ablation_study}, we could conclude that R-HGNN achieves the best performance when it is equipped with all the components and removing any component would lead to worse results. Though the effects of the three components vary on different datasets, all of them contribute to the improvements in the final performance. In particular, the weighted residual connection module combines the target node features with aggregated neighbors' information adaptively, the cross-relation message passing module improves interactions of node representations across different relations, and the relation-aware representations fusing module provides the compact node representation considering semantic representations of relations. 

\subsection{Analysis of Parameter Sensitivity}
\begin{figure}[!htbp]
    \centering
    \includegraphics[width=1.02\columnwidth]{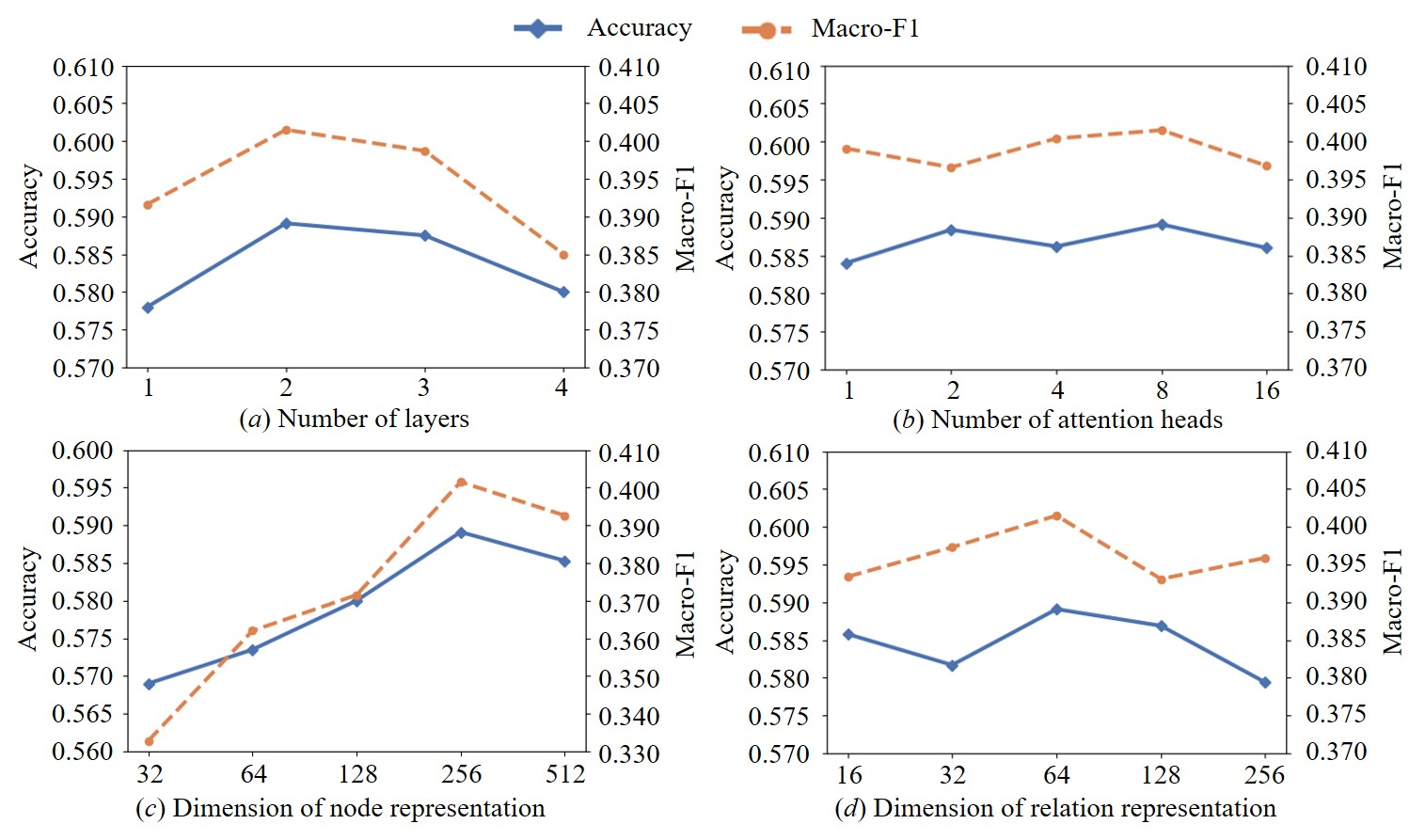}    
    \caption{Parameter Sensitivity of the proposed model on OAG-L1-Field.}
    \label{fig:parameter_sensitivity}
\end{figure}

We study the sensitivity analysis of parameters in R-HGNN, including number of layers and attention heads, and dimension of node and relation representations. Experimental results are reported on the node classification task on OAG-L1-Field in \figref{fig:parameter_sensitivity}.

\textbf{Number of layers}.
We vary the layers in R-HGNN from 1 to 4 and report the result in \figref{fig:parameter_sensitivity} ($a$). We could observe that R-HGNN achieves the best performance when it is stacked with 2 layers. With the increment of layers, the performance of R-HGNN raises at first since deeper architecture allows the model to receive information from multi-hop neighbors. However, the performance starts to drop gradually when too many layers are stacked, which may be caused by the over-smoothing problem \cite{DBLP:conf/aaai/LiHW18}.

\textbf{Number of attention heads}.
We explore the effect of the number of attention heads in R-HGNN and show the result in \figref{fig:parameter_sensitivity} ($b$). It could be concluded that more attention heads would generally improve the model performance and make the training process more stable. When the number of attention heads is set to 8, the performance reaches the top.

\textbf{Dimension of node representation}.
We change the dimension of node representation and report the result in \figref{fig:parameter_sensitivity} ($c$). We could find that the performance of R-HGNN grows with the increment of the node representation dimension and obtains the best performance when the dimension is set to 256. However, the performance decreases when the dimension is larger than 256. Such a phenomenon indicates that an appropriate number of parameters could enhance the model capacity, but too many parameters would lead to the over-parameterization issue and decrease the model generalization ability.

\textbf{Dimension of relation representation}.
Since R-HGNN explicitly captures the semantic representations of relations, we investigate the dimension of relation representation to validate the influence of relations. The result is shown in \figref{fig:parameter_sensitivity} ($d$). We could see that inadvisable dimensions of relation representation decrease the model performance and R-HGNN could achieve satisfactory performance when the dimension of relation representation is properly set to 64.

\subsection{Analysis of Different Relations}
\begin{figure}[!htbp]
    \centering
    \includegraphics[width=1.00\columnwidth]{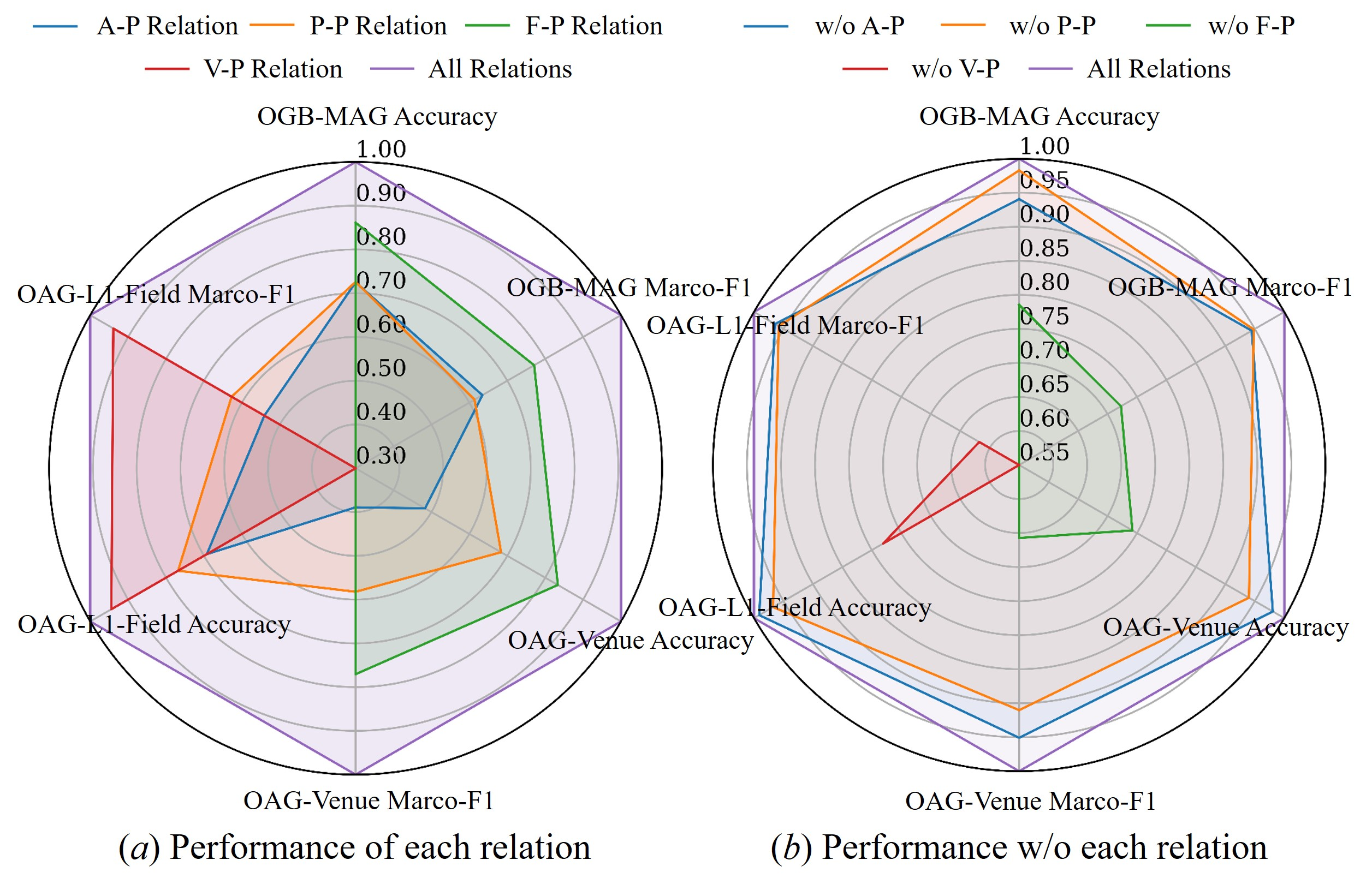}
\caption{Visualization of the performance of different relations.}
\label{fig:relations_performance}
\end{figure}

Since R-HGNN could learn relation-aware node representations in heterogeneous graphs, we aim to investigate on the importance of each relation to the model performance. Specifically, the performance of each relation and the performance without each relation on the three large-scale datasets for the node classification task are presented in \figref{fig:relations_performance}. As the scales of evaluation metrics are different, we report the normalized metrics to make the results more intuitive. It is worth noticing that OGB-MAG and OAG-Venue are not associated with the V-P relation, and OAG-L1-Field does not contain the F-P relation, so their corresponding patterns are not complete as others.  

From \figref{fig:relations_performance}, we could observe that R-HGNN achieves the best performance when considering all the relations, since more relations provide more comprehensive information. Moreover, the importance of different relations varies among different datasets. On the one hand, some specific relations (i.e., the F-P relation on OGB-MAG and OAG-Venue, and the V-P relation on OAG-L1-Field) are essential to the performance and leveraging a single relation could already achieve satisfactory results (see \figref{fig:relations_performance} ($a$)). Excluding the specific relation would lead to the drop in performance drastically (see \figref{fig:relations_performance} ($b$)). On the other hand, some relations are not related to the performance so much. Solely utilizing the specific relation would lead to inferior performance (see \figref{fig:relations_performance} ($a$)) and dropping the specified relation would not significantly affect the model performance (see \figref{fig:relations_performance} ($b$)). 
These findings indicate that it is essential to distinguish the importance of relations and focus on more important relations when learning on heterogeneous graphs.

\subsection{Comparisons of Model Complexity}
\begin{figure}[!htbp]
    \centering
    \includegraphics[width=1.00\columnwidth]{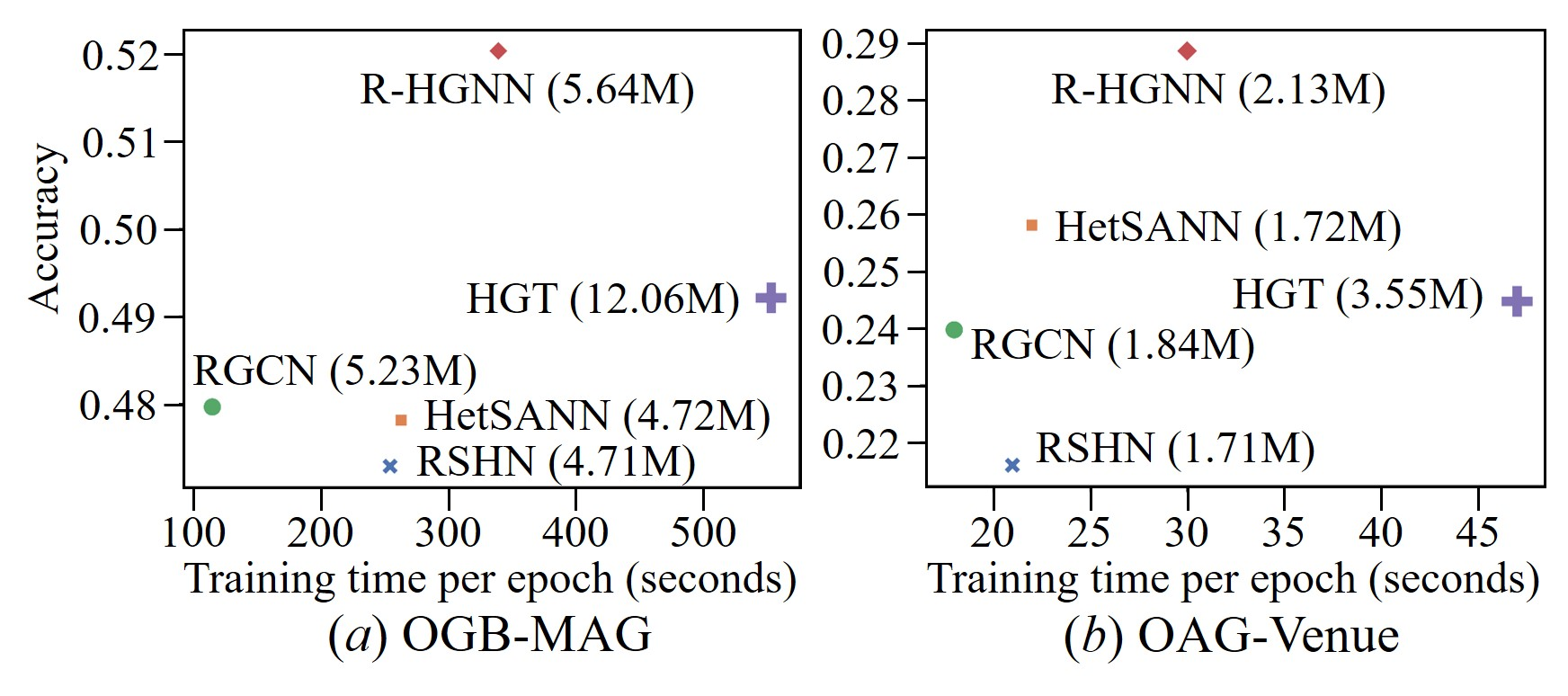}
\caption{Comparisons of the average training time of each epoch and the number of model parameters (shown in parentheses) on OGB-MAG and OAG-Venue datasets. The vertical axis shows classification accuracy.}
\label{fig:model_complexity}
\end{figure}

We also compare the computational cost and parameter capacity of R-HGNN with heterogeneous graph learning baselines on the first two largest datasets, that is, OGB-MAG and OAG-Venue. We conduct the experiments on an Ubuntu machine equipped with one Intel(R) Xeon(R) Gold 6230 CPU @ 2.10GHz with 20 physical cores. The GPU device is NVIDIA Tesla T4 with 15 GB memory. We set the batch size of all the methods to 2560 to make a fair comparison. The average training time of each epoch and the number of model parameters are reported in \figref{fig:model_complexity}.

From \figref{fig:model_complexity}, we could find that R-HGNN achieves a good trade-off between effectiveness and efficiency, which obtains the best performance with a moderate increase in computational cost and parameter capacity compared with the SOTAs. In particular, considering the roles of both nodes and relations helps R-HGNN learn more fine-grained relation-aware node representations and outperform all the SOTAs. Compared with most heterogeneous graph learning methods (e.g., RGCN, RSGN and HetSANN), the increments in training time and parameter capacity are mainly caused by: 1) cross-relation message passing process in \secref{section-3-cross_relation_message_passing}; and 2) relation representation learning process in \secref{section-3-relation_propagate}. The transformer-based attention mechanism in HGT makes it more complicated than other methods, which not only introduces more parameters but also slows down the training speed.

\section{Conclusion}
\label{section-6}

This paper studied the problem of heterogeneous graph learning and proposed to learn node representations considering relation-aware characteristics.
Different from existing research based on the propagation mechanism of node representations, our method exploits the role of relations and collaboratively learns relation-aware node representations as well as semantic representations of relations.
Our method consists of four components: 1) a relation-specific node representation learning module to learn node representations from each relation-specific graph separately; 2) a cross-relation message passing module to facilitate the interactions of node representations across different relations; 3) a relation representation learning module to capture relation semantics; and 4) a semantic fusing module to aggregate relation-aware node representations into a compact representation considering the learned relation representations.
Experimental results on a variety of graph learning tasks demonstrated the superiority of our method over competitive benchmarks. 
This research sheds lights on the way to improve heterogeneous graph learning by exploiting the role of relations. 
In the future work, we will disentangle the relation-aware representations of nodes to improve the interpretability of our approach.


%

%

\ifCLASSOPTIONcompsoc
  \section*{Acknowledgments}
\else
  \section*{Acknowledgment}
\fi
This work was supported in part by the National Natural Science Foundation of China [grant numbers 71901011, 51991391, 51991395, U1811463].

\ifCLASSOPTIONcaptionsoff
  \newpage
\fi



\bibliographystyle{IEEEtran}
\bibliography{reference}

\appendix
\label{section-appendix}
In the appendix, details of the experiments are introduced.

\begin{table*}[!htbp]
\centering
\caption{Settings of dropout and learning rate on all the methods.}
\label{tab:hyperparameters}
\begin{tabular}{c|c|cccccccccc}
\hline
Datasets                      & Hyper-parameters & MLP   & GCN   & GraphSAGE & GAT   & RGCN  & RSHN  & HAN   & HetSANN & HGT   & R-HGNN \\ \hline
\multirow{2}{*}{IMDB}         & dropout          & 0.8   & 0.5   & 0.0       & 0.5   & 0.5   & 0.4   & 0.5   & 0.1     & 0.3   & 0.6     \\ 
                              & learning rate    & 0.005 & 0.01  & 0.005     & 0.001 & 0.005 & 0.001 & 0.005 & 0.01    & 0.01  & 0.001   \\ \hline
\multirow{2}{*}{OGB-MAG}      & dropout          & 0.0   & 0.3   & 0.1       & 0.1   & 0.1   & 0.2   & ---   & 0.3     & 0.1   & 0.5     \\ 
                              & learning rate    & 0.01  & 0.001 & 0.001     & 0.001 & 0.001 & 0.001 & ---   & 0.001   & 0.001 & 0.001   \\ \hline
\multirow{2}{*}{OAG-Venue}    & dropout          & 0.1   & 0.3   & 0.2       & 0.2   & 0.3   & 0.2   & ---   & 0.3     & 0.3   & 0.3     \\
                              & learning rate    & 0.001 & 0.001 & 0.001     & 0.001 & 0.001 & 0.001 & ---   & 0.001   & 0.001 & 0.001   \\ \hline
\multirow{2}{*}{OAG-L1-Field} & dropout          & 0.2   & 0.2   & 0.2       & 0.2   & 0.3   & 0.2   & ---   & 0.2     & 0.1   & 0.3     \\ 
                              & learning rate    & 0.001 & 0.001 & 0.001     & 0.001 & 0.001 & 0.001 & ---   & 0.001   & 0.001 & 0.001   \\ \hline
\end{tabular}
\end{table*}

\subsection*{Details of the Datasets} \label{section-appendix-datasets}
\begin{itemize}
    \item \textbf{IMDB}:
    Plot keywords of movies are provided by the IMDB. Following \cite{DBLP:conf/www/WangJSWYCY19}, we use the bag-of-words representation of plot keywords to denote movie features, corresponding to a 1,537-dimensional feature for each movie. Director/actor features are the average representation of movies that they directed/acted, whose dimensions are both 1,537.
    
    \item \textbf{OGB-MAG}:
    Open Graph Benchmark (OGB) \cite{DBLP:conf/nips/HuFZDRLCL20} contains a diverse set of challenging benchmark datasets for graph machine learning research. Leaderboards are set up for each dataset and state-of-the-art models are ranked based on their performance. Moreover, all the models are listed with open-sourced implementation to reproduce the results. OGB-MAG is a heterogeneous academic network in OGB, where each paper is associated with a 128-dimensional Word2Vec feature. For nodes that do not have features, we generate their features by the metapath2vec \cite{DBLP:conf/kdd/DongCS17} model.
    As a result, the feature of each author/ field/ institution node corresponds to a 128-dimensional vector. The feature of each paper is the concatenation of the given 128-dimensional Word2Vec feature and the generated 128-dimensional structural feature, corresponding to a 256-dimensional vector.
    
    \item \textbf{OAG-Venue}:
    We use the pre-processed graph in the Computer Science (CS) domain extracted from Open Academic Graph (OAG) by \citet{DBLP:conf/www/HuDWS20} to conduct experiments\footnote{HGT authors only shared the graph in the CS domain.}. Features of all types of nodes are given in the OAG dataset. Specifically, the feature of each paper is a 768-dimensional vector, corresponding to the weighted combination of each word's representation in the paper's title. Each word's representation and the attention score are obtained from a pre-trained XLNet \cite{DBLP:conf/nips/YangDYCSL19}. The feature of each author is the average of his/her published paper representations, corresponding to a 768-dimensional vector as well. The features of other types of nodes are generated by the metapath2vec model to reflect the heterogeneous graph structure, whose dimensions are all set to 400. One potential issue with the OAG dataset in \cite{DBLP:conf/www/HuDWS20} is the information leakage, since target nodes and the nodes with ground truth are connected with edges. To solve this issue, we remove all the edges between paper nodes and nodes with ground truth that we aim to predict. Specifically, the classification task on OAG-Venue is to predict the published venues of papers, so we remove all edges between paper nodes and venue nodes in the original OAG dataset. We select venues that associated with no less than 200 papers to conduct experiments. In total, there are 241 venues in OAG-Venue, making the task as a 241-class classification problem.
    
    \item \textbf{OAG-L1-Field}:
    The classification task on OAG-L1-Field is to predict the $L1$-level field that each paper belongs to, so we remove all the edges between paper nodes and field nodes in the original OAG dataset. We select fields that associated with no less than 100 papers to conduct experiments. In total, there are 52 fields in OAG-L1-Field, making the task as a 52-class classification problem. 
\end{itemize}

\subsection*{Selection of Dropout and Learning Rate} \label{section-appendix-hyper-parameters}
On IMDB, the dropout and learning rate are searched in $\left[0.0, 0.1, \cdots, 0.9\right]$ and $\left[0.001, 0.005, 0.01\right]$, respectively. On OGB-MAG, we search the dropout and learning rate in $\left[0.0, 0.1, 0.2, 0,3, 0.4, 0.5\right]$ and $\left[0.001, 0.01\right]$. On OAG-Venue and OAG-L1-Field,the dropout and learning rate are searched in $\left[0.0, 0.1, 0.2, 0,3\right]$ and $\left[0.001, 0.01\right]$, respectively. The settings of dropout and learning rate on all the methods are shown in \tabref{tab:hyperparameters}.


\subsection*{Node Clustering} \label{section-appendix-node_clustering}
On the small-scale dataset, we feed the learned representations of all the movie nodes into k-means algorithm to achieve the clustering performance of different models. On large-scale datasets, it is infeasible to feed all the paper nodes into k-means algorithm. Therefore, we first select top-five classes of papers in the testing set and then randomly select 1000 papers from each class, and finally obtain 5,000 papers. Then we feed the selected 5,000 paper nodes into k-means algorithm to get the clustering results. The number of clusters is equal to the number of real classes in each dataset (i.e., 3 for IMDB, and 5 for OGB-MAG, OAG-Venue and OAG-L1-Field).

\subsection*{Link Prediction} \label{section-appendix-link_prediction}
Due to the huge number of edges on large-scale datasets, it is infeasible to do link prediction on all the edges. Therefore, we adjust the number of sampled edges on the datasets. In particular, 3\%, 1\% and 1\% of the edges are sampled as training, validation and testing sets on OGB-MAG, respectively. Correspondingly, 15\%, 5\% and 5\% on OAG-Venue, and 30\%, 10\% and 10\% on OAG-L1-Field. Each edge in the training set is associated with five randomly sampled negative edges, and each edge in the validation or testing sets is associated with a randomly sampled negative edge.

%


\begin{IEEEbiography}[{\includegraphics[width=1in,height=1.25in,clip,keepaspectratio]{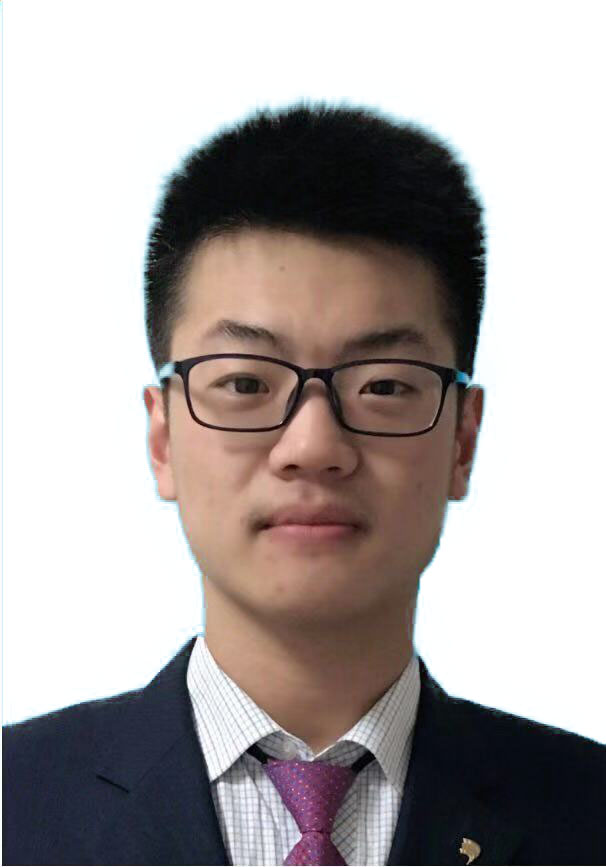}}]
{Le Yu} received the B.S. degree in Computer Science and Engineering from Beihang University, Beijing, China, in 2019. He is currently a third-year computer science Ph.D. student in the School of Computer Science and Engineering at Beihang University. His research interests include representation learning, graph neural networks and temporal data mining.
\end{IEEEbiography}

\begin{IEEEbiography}[{\includegraphics[width=1in,height=1.25in,clip,keepaspectratio]{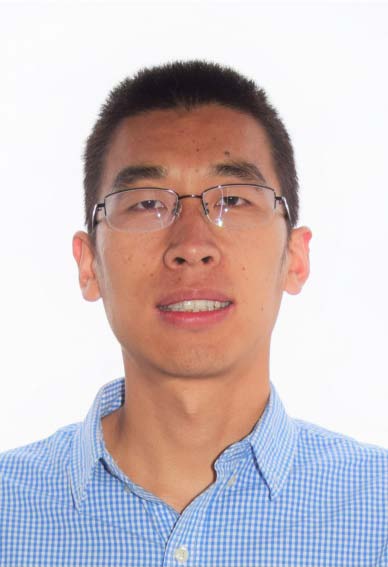}}]
{Leilei Sun} is currently an assistant professor in School of Computer Science, Beihang University, Beijing, China. He was a postdoctoral research fellow from 2017 to 2019 in School of Economics and Management, Tsinghua University. He received his Ph.D. degree from Institute of Systems Engineering, Dalian University of Technology, in 2017. His research interests include machine learning and data mining. 
\end{IEEEbiography}

\begin{IEEEbiography}[{\includegraphics[width=1in,height=1.25in,clip,keepaspectratio]{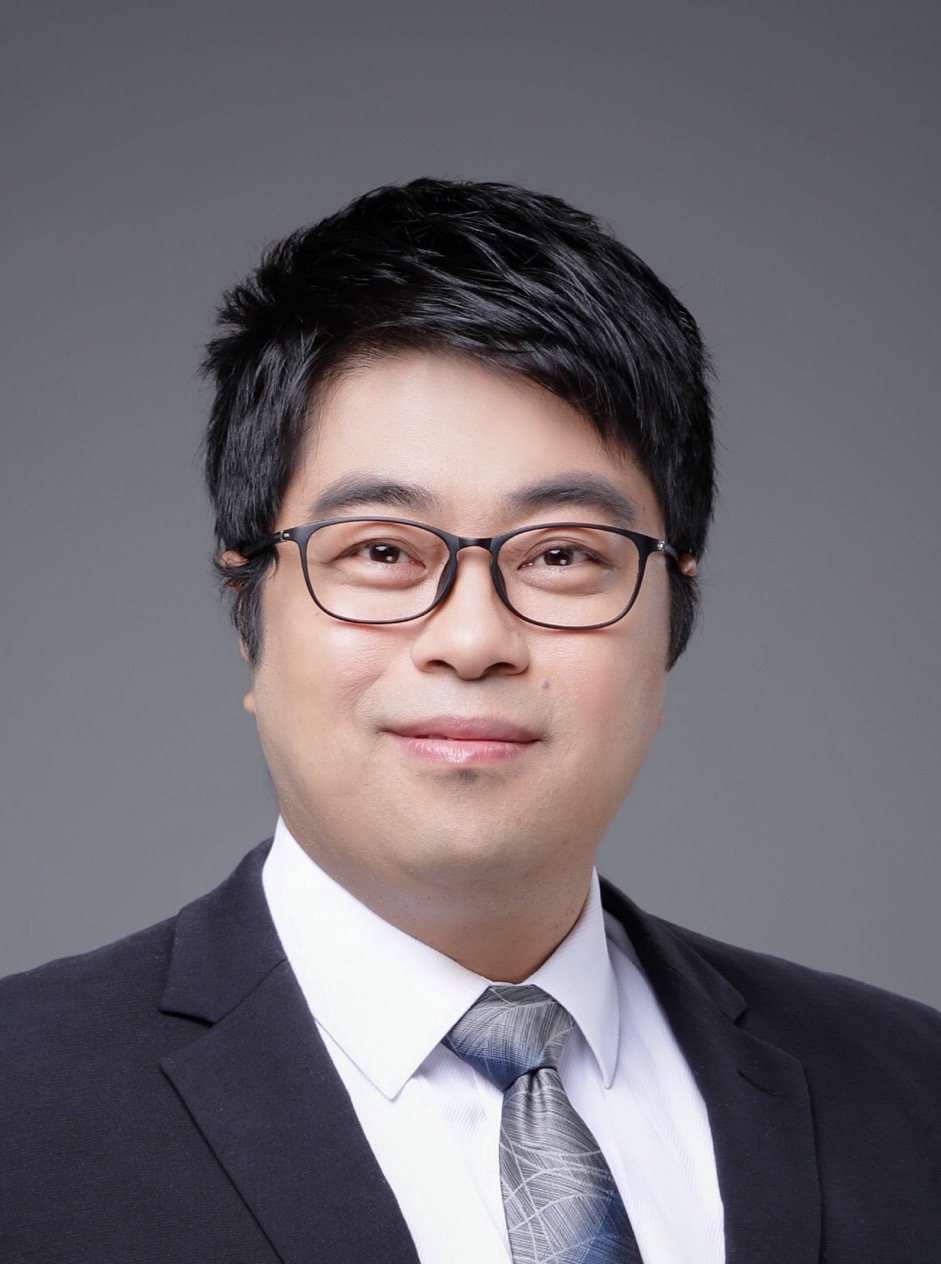}}]
{Bowen Du} received the Ph.D. degree in Computer Science and Engineering from Beihang University, Beijing, China, in 2013. He is currently a Professor with the State Key Laboratory of Software Development Environment, Beihang University. His research interests include smart city technology, multi-source data fusion, and traffic data mining.
\end{IEEEbiography}

\begin{IEEEbiography}[{\includegraphics[width=1in,height=1.25in,clip,keepaspectratio]{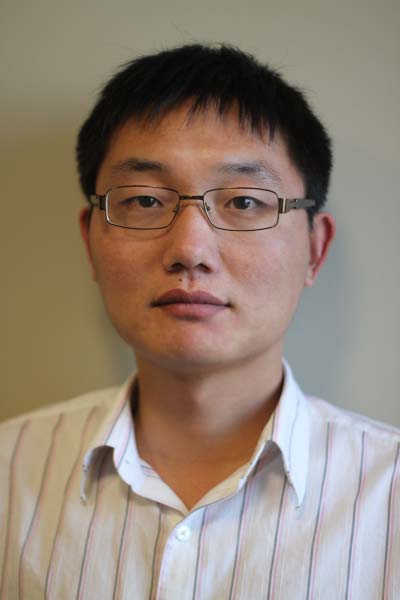}}]
{Chuanren Liu} received the B.S. degree from the University of Science and Technology of China (USTC), the M.S. degree from the Beijing University of Aeronautics and Astronautics (BUAA), and the Ph.D. degree from Rutgers, the State University of New Jersey.
He is currently an assistant professor with the Business Analytics and Statistics Department at the University of Tennessee, Knoxville, USA.
His research interests include data mining and machine learning, and their applications in business analytics.
\end{IEEEbiography}

\newpage

\begin{IEEEbiography}[{\includegraphics[width=1in,height=1.25in,clip,keepaspectratio]{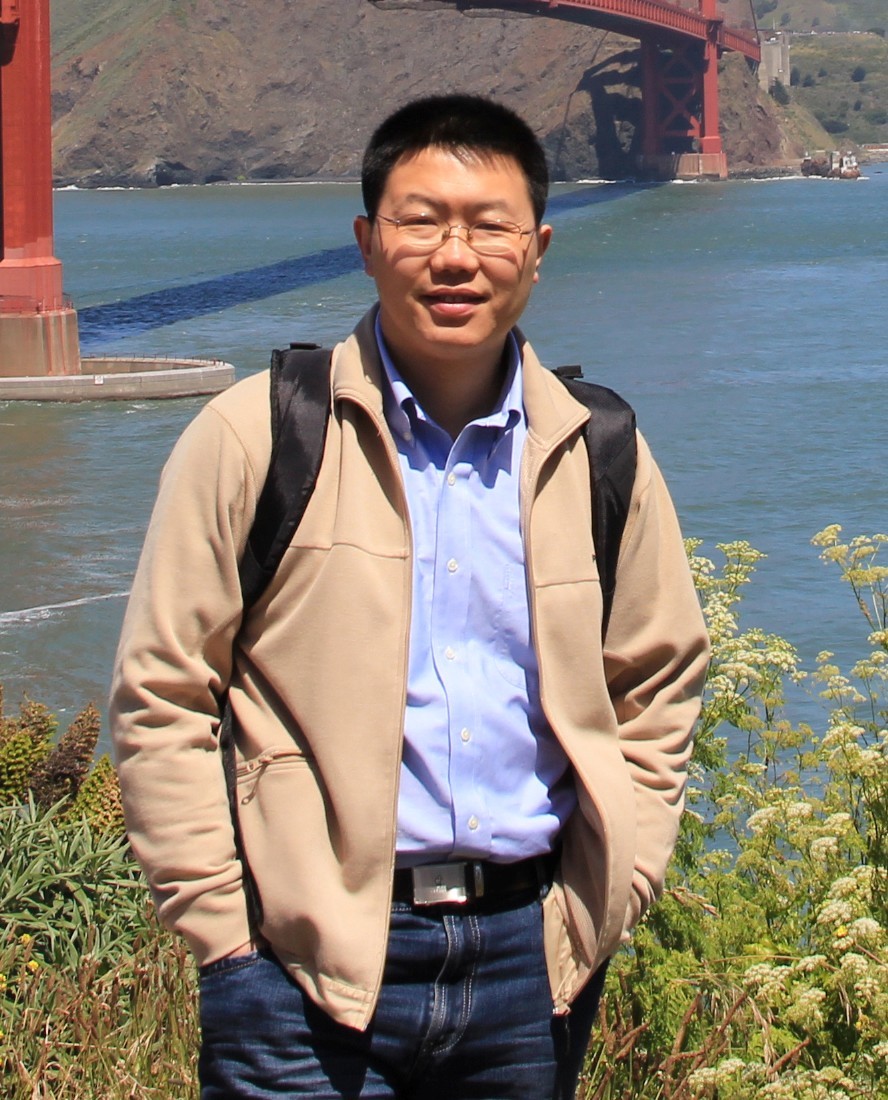}}]
{Weifeng Lv} received the B.S. degree in Computer Science and Engineering from Shandong University, Jinan, China, and the Ph.D. degree in Computer Science and Engineering from Beihang University, Beijing, China, in 1992 and 1998 respectively. Currently, he is a Professor with the State Key Laboratory of Software Development Environment, Beihang University, Beijing, China. His research interests include smart city technology and mass data processing.
\end{IEEEbiography}

\begin{IEEEbiography}[{\includegraphics[width=1in,height=1.25in,clip,keepaspectratio]{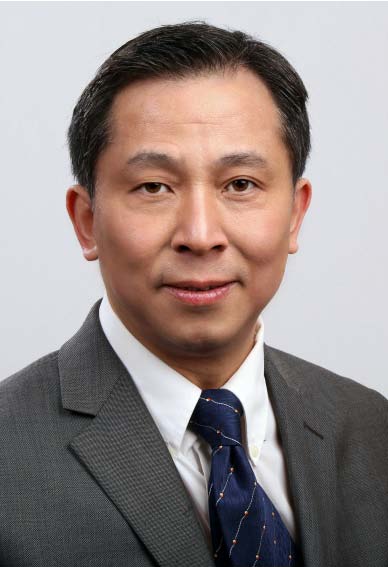}}]
{Hui Xiong} is currently a Chair Professor at the Hong Kong University of Science and Technology (Guangzhou). Dr. Xiong’s research interests include data mining, mobile computing, and their applications in business. Dr. Xiong received his PhD in Computer Science from University of Minnesota, USA. He has served regularly on the organization and program committees of numerous conferences, including as a Program Co-Chair of the Industrial and Government Track for the 18th ACM SIGKDD International Conference on Knowledge Discovery and Data Mining (KDD), a Program Co-Chair for the IEEE 2013 International Conference on Data Mining (ICDM), a General Co-Chair for the 2015 IEEE International Conference on Data Mining (ICDM), and a Program Co-Chair of the Research Track for the 2018 ACM SIGKDD International Conference on Knowledge Discovery and Data Mining. He received the 2021 AAAI Best Paper Award and the 2011 IEEE ICDM Best Research Paper award. For his outstanding contributions to data mining and mobile computing, he was elected an AAAS Fellow and an IEEE Fellow in 2020.
\end{IEEEbiography}




\end{document}